\documentclass[11pt, a4paper, logo, copyright]{googledeepmind}

\usepackage[authoryear, sort&compress, round]{natbib}
\usepackage[capitalise]{cleveref}
\usepackage{amsfonts}
\usepackage{titlesec}
\usepackage{amsmath}
\usepackage{graphicx}
\usepackage{paralist}
\usepackage{listings}
\usepackage{enumitem}
\usepackage{authblk}
\usepackage{ifthen}
\usepackage{defs}
\usepackage{float}
\usepackage{subcaption}
\newboolean{shownotes}
\setboolean{shownotes}{false}
\newcommand{\rohin}[1]{\ifthenelse{\boolean{shownotes}}{{\color{blue} RS: #1}}{}
}

\newcommand{\todo}[2][]{\textcolor{red}{\textbf{TODO\ifthenelse{ \equal{#1}{} }{}{(#1)}:} #2}}
\newcommand{\TODO}[2][]{\textcolor{red}{\textbf{TODO\ifthenelse{ \equal{#1}{} }{}{(#1)}:} #2}}
\newcommand{\norm}[1]{\left\lVert #1 \right\rVert}

\title{Improving Dictionary Learning with Gated Sparse Autoencoders}

\correspondingauthor{srajamanoharan@google.com and neelnanda@google.com}

\author[*]{Senthooran Rajamanoharan}
\author[*]{Arthur Conmy}
\author[ \hspace{-0.7ex}]{Lewis Smith}
\author[ \hspace{-0.7ex}]{Tom Lieberum$^\dagger$}
\author[ \hspace{-0.7ex}]{Vikrant Varma$^\dagger$}
\author[ \hspace{-0.7ex}]{J\'anos Kram\'ar}
\author[ \hspace{-0.7ex}]{Rohin Shah}
\author[ \hspace{-0.7ex}]{Neel Nanda}

\affil[*]{: Joint contribution. $^\dagger$: Core infrastructure contributor.}

\begin{abstract}

Recent work has found that sparse autoencoders (SAEs) are an effective technique for unsupervised discovery of interpretable features in language models' (LMs) activations, by finding sparse, linear reconstructions of LM activations.
We introduce the Gated Sparse Autoencoder (Gated SAE), which achieves a Pareto improvement over training with prevailing methods.
In SAEs, the L1 penalty used to encourage sparsity introduces many undesirable biases, such as \emph{shrinkage} -- systematic underestimation of feature activations.
The key insight of Gated SAEs is to separate the functionality of (a) determining which directions to use and (b) estimating the magnitudes of those directions: this enables us to apply the L1 penalty only to the former, limiting the scope of undesirable side effects.
Through training SAEs on LMs of up to 7B parameters we find that, in typical hyper-parameter ranges, Gated SAEs solve shrinkage, are similarly interpretable, and require half as many firing features to achieve comparable reconstruction fidelity.

\end{abstract}
\begin{document}
\maketitle
\section{Introduction}
\label{sec:intro}

Mechanistic interpretability research aims to explain how neural networks produce outputs in terms of the learned algorithms executed during a forward pass \citep{AnthropicMechanisticEssay, olah2020zoom}. Much work makes use of the fact that many concept representations appear to be linear \citep{elhage2021mathematical, gurnee2023finding, olah2020zoom, park2023linear}. However, finding the set of all interpretable directions is a highly non-trivial problem. Classic approaches, like interpreting neurons (i.e. directions in the standard basis) are insufficient, as many are \emph{polysemantic} and tend to activate for a range of different seemingly unrelated concepts~\citep[Empirical Phenomena]{bolukbasi2021interpretability,elhage2022solu,elhage2022toy}. %

The \emph{superposition hypothesis}~\citep[Definitions and Motivation]{elhage2022toy} posits a mechanistic explanation for these observations: in an intermediate representation of dimension $\dAct$, a model will encode $\dFeat \gg \dAct$ concepts as linear directions, where the set of concepts and their directions is fixed across all inputs, but on a given input only a sparse subset of concepts are active, ensuring that there is not much simultaneous interference \citep[Appendix A]{gurnee2023finding} between these (non-orthogonal) concepts. Motivated by the superposition hypothesis, \citet{bricken2023monosemanticity} and \citet{cunningham2023sparse} recently used sparse autoencoders (SAEs; \citet{ng2011sparse}) to find sparse decompositions of model activations in terms of an overcomplete basis, or \emph{dictionary} \citep{mallat1993matchingpursuits}.\footnote{Although motivated by the superposition hypothesis, the utility of line of research is not contingent on this hypothesis being true. If a faithful, sparse and interpretable decomposition can be found, we expect this to be a useful basis in its own right for downstream interpretability tasks, such as understanding or intervening on a model's representations and circuits, even if some fraction of the model's computation is e.g.~represented non-linearly and not captured.} 

Although SAEs show promise in this regard, the L1 penalty used in the prevailing training method to encourage sparsity also introduces biases that harm the accuracy of SAE reconstructions, as the loss can be decreased by trading-off some reconstruction accuracy for lower L1. We refer to this existing training methodology as the \emph{baseline SAE}, defined fully in \Cref{subsec:background_arch}-\ref{subsec:background_training} and which borrows heavily from \citet{bricken2023monosemanticity}. In this paper, we introduce a modification to the baseline SAE architecture -- a \emph{Gated SAE} -- along with an accompanying loss function, which partially overcomes these limitations. Our key insight is to use separate affine transformations for (a) determining which dictionary elements to use in a reconstruction and (b) estimating the coefficients of active elements, and to apply the sparsity penalty only to the former task. We share a subset of weights between these transformations to avoid significantly increasing the parameter count and inference-time compute requirements of a Gated SAE compared to a baseline SAE of equivalent width.\footnote{Although due to an auxiliary loss term, computing the Gated SAE loss for training purposes does require~50\% more compute than computing the loss for a matched-width baseline SAE.}

\begin{figure}[t]
\centering
\includegraphics[width=\textwidth]{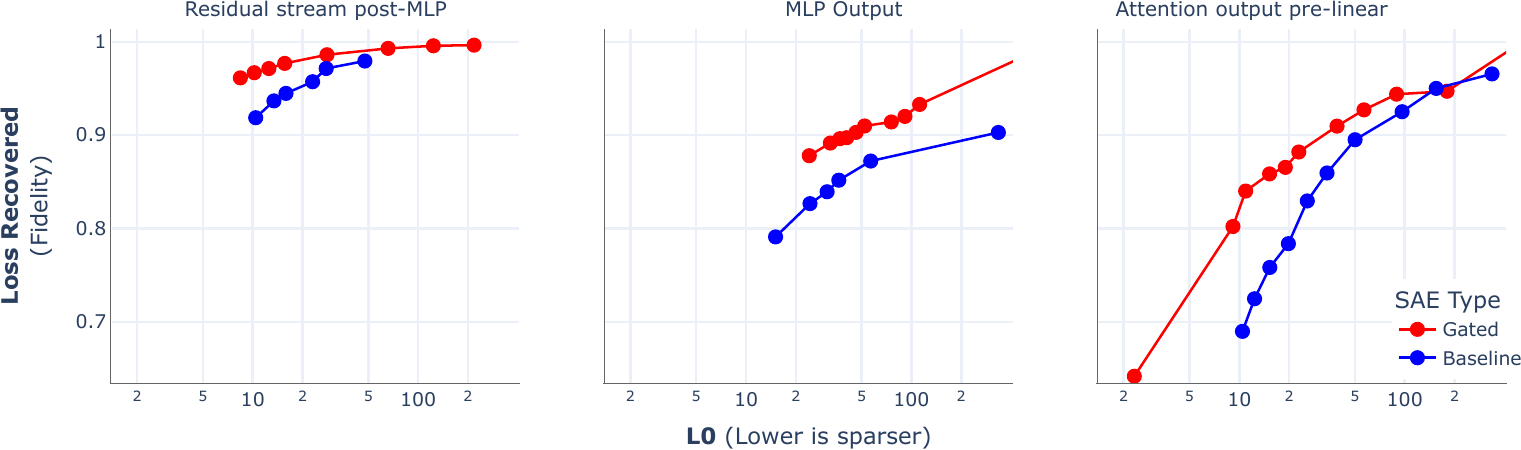}
\caption{The performance of Gated SAEs compared to the baseline SAE at Layer 20 in Gemma-7B (log-scale axes from L0=2 to L0=200). The SAEs are trained with equal compute, since the baseline SAEs have 50\% more learned features (\Cref{subsec:benchmarking}). This performance improvement holds in layers throughout GELU-1L, Pythia-2.8B and Gemma-7B (\Cref{app:more_paretos}). Full detail in \Cref{table:gemma_7b_baselines2} and \ref{table:gemma_7b_gated2}.}
\label{fig:headline_pareto_plot}
\end{figure}

We evaluate Gated SAEs on multiple models: a one layer GELU activation language models \citep{gelu1l}, Pythia-2.8B \citep{biderman2023pythia} and Gemma-7B \citep{gemma_2024}, and on multiple sites within models: MLP layer outputs, attention layer outputs, and residual stream activations. Across these models and sites, we find Gated SAEs to be a Pareto improvement over baseline SAEs holding training compute fixed (\cref{fig:headline_pareto_plot}): they yield sparser decompositions at any desired level of reconstruction fidelity. We also conduct further follow up ablations and investigations on a subset of these models and sites to better understand the differences between Gated SAEs and baseline SAEs.

Overall, the key contributions of this work are that we:
\begin{compactenum}
\item Introduce the Gated SAE, a modification to the standard SAE architecture that decouples detection of which features are present from estimating their magnitudes (\Cref{subsec:defining_gated_saes});
\item Show that Gated SAEs Pareto improve the sparsity and reconstruction fidelity trade-off, compared to baseline SAEs (\Cref{subsec:benchmarking});
\item Confirm that Gated SAEs overcome the shrinkage problem (\Cref{subsec:shrinkage}), while outperforming other methods that also address this problem (\Cref{subsec:ablation_study});
\item Provide evidence from a small double-blind study that Gated SAE features are comparably interpretable to baseline SAE features (\Cref{subsec:manual_interp}).
\end{compactenum}

\section{Sparse Autoencoder Background}
\label{sec:background}

In this section we summarise the concepts and notation necessary to understand existing SAE architectures and training methods, which we call the \emph{baseline SAE}. We define Gated SAEs in \Cref{subsec:defining_gated_saes}. We follow notation broadly similar to \citet{bricken2023monosemanticity} and recommend that work as a more complete introduction to training SAEs on LMs.  

As motivated in \Cref{sec:intro}, we wish to decompose a model's activation $\mathbf{x} \in \mathbb{R}^\dAct$ into a sparse, linear combination of feature directions:
\begin{equation}
\mathbf{x} \approx \mathbf{x}_0 + \sum_{i=1}^{\dFeat} f_i(\mathbf{x}) \mathbf{d}_i,
\label{eqn:superposition-decomposition}
\end{equation}
where $\mathbf{d}_i$ are $\dFeat\gg\dAct$ latent unit-norm \emph{feature directions}, and the sparse coefficients $f_i(\mathbf{x}) \geq 0$ are the corresponding \emph{feature activations} for $\mathbf{x}$.\footnote{In this work, we use the term \textit{feature} only in the context of the \textit{learned features} of SAEs, i.e. the overcomplete basis directions that are linearly combined to produce reconstructions. In particular, \textit{learned features} are always linear and not necessarily interpretable, sidestepping the difficulty in defining what a feature is (\citet{elhage2022toy}'s `What are features?' section).} The right-hand side of \cref{eqn:superposition-decomposition} naturally has the structure of an autoencoder: an input activation $\mathbf{x}$ is encoded into a (sparse) feature activations vector $\mathbf{f}(\mathbf{x})\in \mathbb{R}^\dFeat$, which in turn is linearly decoded to reconstruct $\mathbf{x}$.

\subsection{Baseline Architecture}
\label{subsec:background_arch}

Using this correspondence, \citet{bricken2023monosemanticity} and subsequent works attempt to learn a suitable sparse decomposition by parameterising a single-layer autoencoder $\left(\mathbf{f}, \hat{\mathbf{x}}\right)$ defined by:
\begin{align}
    {\mathbf{f}}(\mathbf{x}) &:= \text{ReLU} \left(\Wenc\left(\mathbf{x} - \bdec\right) + \benc \right) \label{eqn:encoder}\\
    \hat{\mathbf{x}}({\mathbf{f}}) &:= \Wdec {\mathbf{f}} + \bdec
    \label{eqn:decoder}
\end{align}
and training it (\Cref{subsec:background_training}) to reconstruct a large dataset of model activations $\mathbf{x}\sim\mathcal{D}$, constraining the hidden representation $\mathbf{f}$ to be sparse.\footnote{Model activations are typically taken from a specific layer and site, e.g.~the output of the MLP part of layer 17.} Once the sparse autoencoder has been trained, we obtain a decomposition of the form of \cref{eqn:superposition-decomposition} by identifying the (suitably normalised) columns of the decoder weight matrix $\Wdec \in \mathbb{R}^{\dFeat\times\dAct}$ with the dictionary of feature directions $\mathbf{d_i}$, the decoder bias $\bdec \in \mathbb{R}^{\dAct}$ with the centering term $\mathbf{x}_0$, and the (suitably normalised) entries of the latent representation $\mathbf{f}(\mathbf{x}) \in \mathbb{R}^\dFeat$ with the feature activations $f_i(\mathbf{x})$.

\subsection{Baseline Training Methodology}
\label{subsec:background_training}

To train sparse autoencoders, \citet{bricken2023monosemanticity} use a loss function that jointly encourages (i) faithful reconstruction and (ii) sparsity. Reconstruction fidelity is encouraged by the squared distance between SAE input and its reconstruction, $\norm{\mathbf{x}-\hat{\mathbf{x}}(\mathbf{f}(\mathbf{x}))}_2^2$, which we call the \emph{reconstruction loss}, whereas sparsity is encouraged by the L1 norm of the active features, $\norm{\mathbf{f}(\mathbf{x})}_1$, which we call the \emph{sparsity penalty}.\footnote{Note that we cannot directly optimize the L0 norm (i.e. the number of active features) since this is not a differentiable function. We do however use the L0 norm to evaluate SAE sparsity (\Cref{sec:evaluation}).} Balancing these two terms with a \textit{L1 coefficient} $\lambda$, the loss used to optimize SAEs is given by 
\begin{equation}
    \mathcal{L}(\mathbf{x}) := \norm{\mathbf{x}-\hat{\mathbf{x}}(\mathbf{f}(\mathbf{x}))}_2^2 + \lambda \norm{\mathbf{f}(\mathbf{x})}_1.
    \label{eqn:sae_loss}
\end{equation}
Since it is possible to arbitrarily reduce the sparsity loss term without affecting reconstructions or sparsity by simply scaling down encoder outputs and scaling up the norm of the decoder weights, it is important to constrain the norms of the columns of $\Wdec$ during training. Following \citet{bricken2023monosemanticity}, we constrain columns to have exactly unit norm. See \cref{app:training_hyperparameters_and_details} for full details about our (Gated and baseline) SAE training.

\subsection{Evaluation}
\label{subsec:background_evaluation}

To get a sense of the quality of trained SAEs we use two metrics from \citet{bricken2023monosemanticity}: \textbf{L0}, a measure of SAE sparsity and \textbf{loss recovered}, a measure of SAE reconstruction fidelity.

\begin{compactitem}
\item The \textbf{L0} of a SAE is defined by the average number of active features on a given input, i.e $\mathbb{E}_{\mathbf{x} \sim \mathcal{D}} \norm{\mathbf{f}(\mathbf{x})}_0$.
\item The \textbf{loss recovered} of a SAE is calculated from the average cross-entropy loss of the language model on an evaluation dataset, when the SAE's reconstructions are spliced into it. If we denote by $\text{CE}(\mathbf{\phi})$ the average loss of the language model when we splice in a function $\mathbf{\phi}: \mathbb{R}^\dAct \to \mathbb{R}^\dAct$ at the SAE's site during the model's forward pass, then loss recovered is 
\begin{equation}
    1 - \frac{ \text{CE}(\mathbf{\hat{x}\circ\mathbf{f}}) - \text{CE}(\mathbf{\text{Id}})}
    {\text{CE}(\zeta) - \text{CE}(\mathbf{\text{Id}})},
    \label{eqn:loss_recovered}
\end{equation}
Where $\mathbf{\hat{x}\circ\mathbf{f}}$ is the autoencoder function, $\zeta: \mathbf{x}\mapsto \mathbf{0}$ the zero-ablation function and $\text{Id}:\mathbf{x}\mapsto\mathbf{x}$ the identity function. According to this definition, a SAE that always outputs the zero vector as its reconstruction would get a loss recovered of 0\%, whereas a SAE that reconstructs its inputs perfectly would get a loss recovered of 100\%.
\end{compactitem}
Of course, these metrics do not paint the full picture of SAE quality,\footnote{For example, see \citet[Tanh Penalty in Dictionary Learning]{anthropic_february_update}.} hence we perform manual analysis of SAE interpretability in \Cref{subsec:manual_interp}.

\section{Gated SAEs}

\subsection{Motivation}
\label{subsec:motivation}

\begin{figure}[t]
\begin{minipage}[b]{0.65\textwidth} 
\centering
\includegraphics[width=\linewidth]{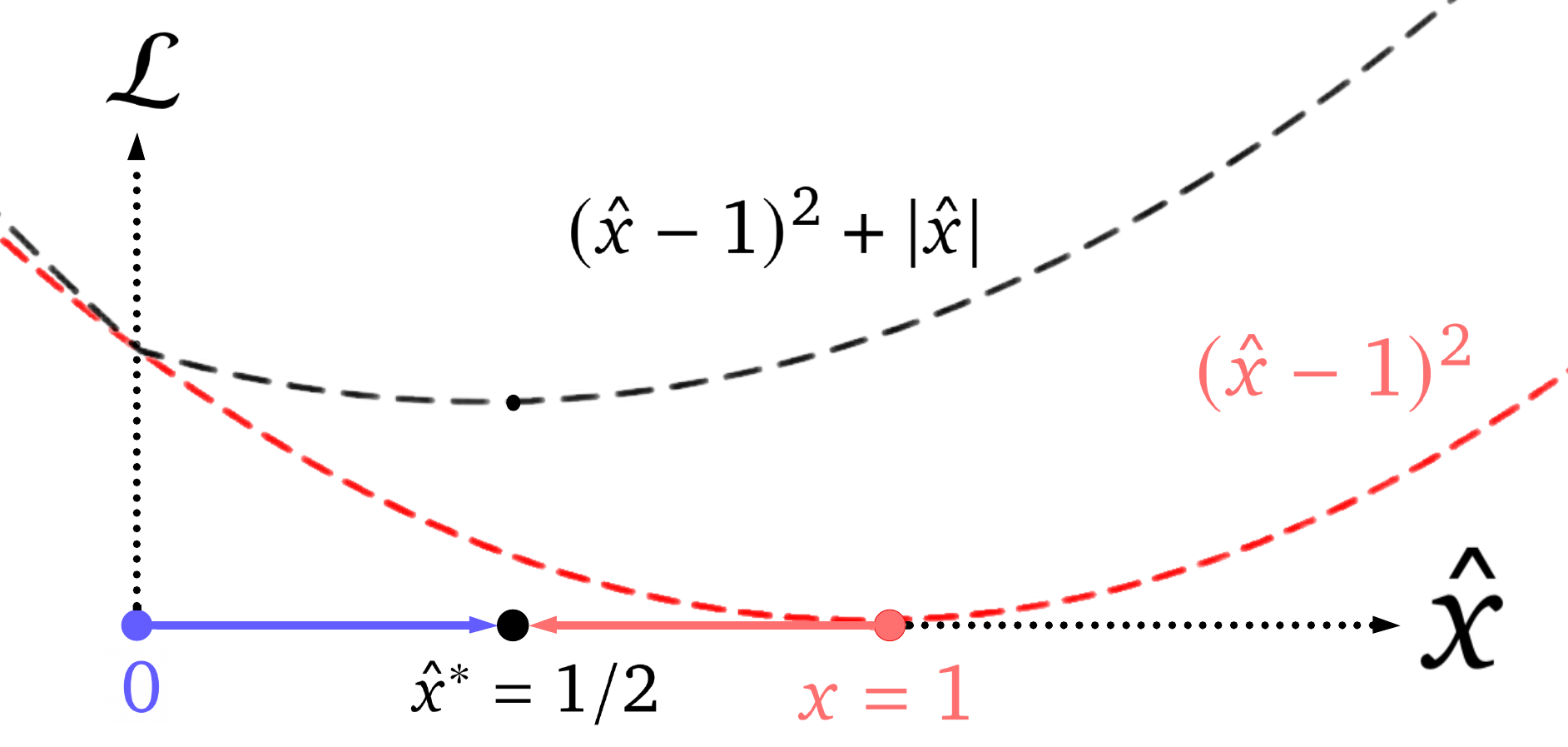}
\end{minipage}
\hfill
\begin{minipage}[b]{0.33\textwidth}
\caption{ \raggedright The L1 penalty in sparse autoencoder causes \textit{shrinkage} -- reconstructions are biased towards smaller norms, even when perfect reconstruction is possible. \\
E.g. a single-feature SAE (with L1 coefficient $\lambda=1$) reconstructs 1/2 rather than 1 when minimizing \Cref{eqn:sae_loss}.}
\label{fig:shrinkage}
\end{minipage}
\end{figure}

The intuition behind how SAEs are trained is to maximise reconstruction fidelity at a given level of sparsity, as measured by L0, although in practice we optimize a mixture of reconstruction fidelity and L1 regularization. This difference is a source of unwanted bias in the training of a sparse autoencoder: for any fixed level of sparsity, a trained SAE can achieve lower loss (as defined in \cref{eqn:sae_loss}) by trading off a little reconstruction fidelity to perform better on the L1 sparsity penalty.

The clearest consequence of this bias is \emph{shrinkage} \citep{wright2024addressing}, illustrated in \Cref{fig:shrinkage}. Holding the decoder $\hat{\mathbf{x}}(\bullet)$ fixed, the L1 penalty pushes feature activations $\mathbf{f}(\mathbf{x})$ towards zero, while the reconstruction loss pushes $\mathbf{f}(\mathbf{x})$ high enough to produce an accurate reconstruction. Thus, the optimal value is somewhere in between, which means it systematically underestimates the magnitude of feature activations, without any necessarily having any compensatory benefit for sparsity.\footnote{Conversely, rescaling the shrunk feature activations \citep{wright2024addressing} is not necessarily enough to overcome the bias induced by by L1 penalty: a SAE trained with the L1 penalty could have learnt sub-optimal encoder and decoder directions that are not improved by such a fix. In \Cref{subsec:rescale_expt} and \Cref{fig:pareto_ito} we provide empirical evidence that this is true in practice.}

How can we reduce the bias introduced by the L1 penalty? The output of the encoder $\mathbf{f}(\mathbf{x})$ of a baseline SAE (\Cref{subsec:background_arch}) has two roles:
\begin{compactenum}
\item It \emph{detects} which features are active (according to whether the outputs are zero or strictly positive). For this role, the L1 penalty is necessary to ensure the decomposition is sparse.
\item It \emph{estimates} the magnitudes of active features. For this role, the L1 penalty is a source of unwanted bias.
\end{compactenum}
If we could separate out these two functions of the SAE encoder, we could design a training loss that narrows down the scope of SAE parameters that are affected (and therefore to some extent biased) by the L1 sparsity penalty to precisely those parameters that are involved in feature detection, minimising its impact on parameters used in feature magnitude estimation.

\subsection{Gated SAEs}
\label{subsec:defining_gated_saes}

\subsubsection{Architecture}

How should we modify the baseline SAE encoder to achieve this separation of concerns? Our solution is to replace the single-layer ReLU encoder of a baseline SAE with a \emph{gated} ReLU encoder. Taking inspiration from Gated Linear Units \citep{shazeer2020glu, dauphin2017gated}, we define the gated encoder as follows:
\begin{equation}
    \tilde{\mathbf{f}}(\mathbf{x}) := \underbrace{\mathbb{1}[\overbrace{(\Wgate(\mathbf{x} - \bdec) + \bgate)}^{\pigate(\mathbf{x})} > \mathbf{0}]}_{\fgate(\mathbf{x})} \odot \underbrace{\text{ReLU}(\Wmag(\mathbf{x} - \bdec) + \bmag)}_{\fmag(\mathbf{x})},
    \label{eq:gated_encoder}
\end{equation}
where $\mathbb{1}[\bullet>\mathbf0]$ is the (pointwise) Heaviside step function and $\odot$ denotes elementwise multiplication. Here, $\fgate$ determines which features are deemed to be active, while $\fmag$ estimates feature activation magnitudes (which only matter for features that have been deemed to be active); $\pigate(\mathbf{x})$ are the $\fgate$ sub-layer's pre-activations, which are used in the gated SAE loss, defined below.

\begin{figure}[t]
\centering
\includegraphics[width=\textwidth]{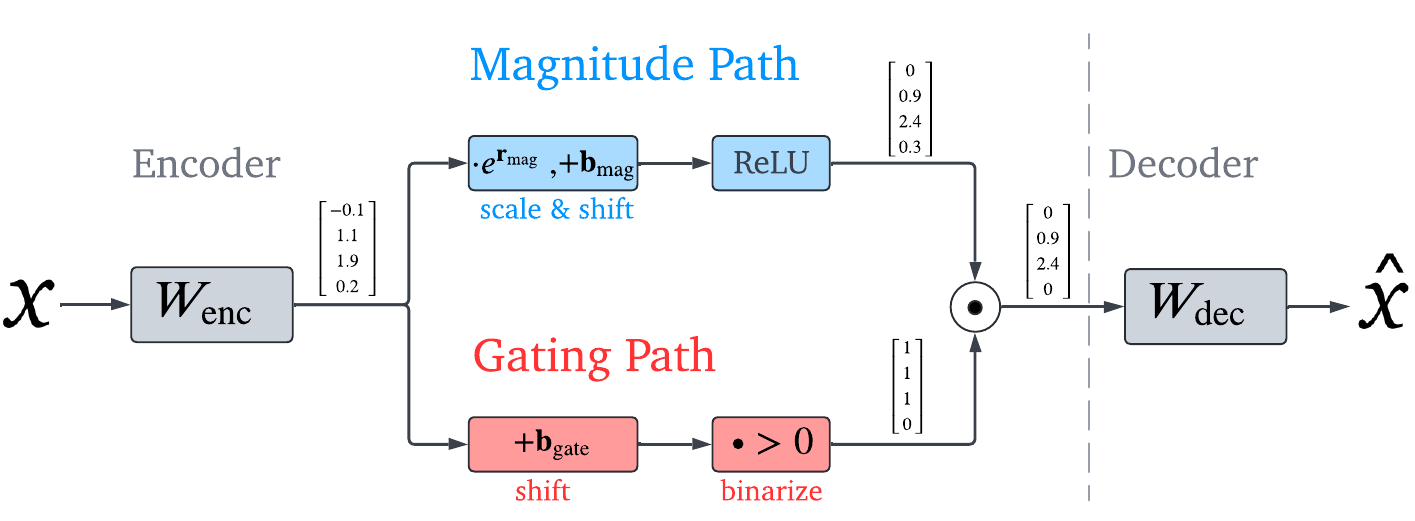}
\caption{The Gated SAE architecture with weight sharing between the gating and magnitude paths, shown with an example input.}
\label{fig:gated_sae_architecture}
\end{figure}

Naively, we appear to have doubled the number of parameters in the encoder, increasing the total number of parameters by 50\%. We mitigate this through weight sharing: we parameterise these layers so that the two layers share the same projection directions, but allow the norms of these directions as well as the layer biases to differ. Concretely, we define $\Wmag$ in terms of $\Wgate$ and an additional vector-valued rescaling parameter $\rmag \in \mathbb{R}^\dFeat$ as follows:
\begin{equation}
    \left(\Wmag\right)_{ij} :=
    \left(\exp(\rmag)\right)_{i} \cdot \left(\Wgate\right)_{ij}.
\label{eq:tying_scheme}
\end{equation}
See \cref{fig:gated_sae_architecture} for an illustration of the tied-weight Gated SAEs architecture.
With this weight tying scheme, the Gated SAE has only $2\times\dFeat$ more parameters than a baseline SAE.
In \cref{subsec:ablation_study}, we perform an ablation study showing that this weight tying scheme leads to a small increase in performance.

\begin{figure}[t]
\centering
\includegraphics[width=0.33\textwidth]{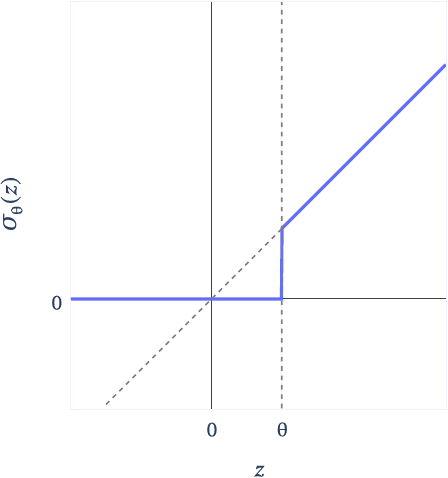}
\caption{After applying the weight sharing scheme of \cref{eq:tying_scheme}, a gated encoder becomes equivalent to a single layer linear encoder with a Jump ReLU \citep{erichson2019jumprelu} activation function $\sigma_\theta$, illustrated above.}
\label{fig:jump_relu}
\end{figure}

With tied weights, the gated encoder can be reinterpreted as a single-layer linear encoder with a non-standard and discontinuous ``Jump ReLU'' activation function \citep{erichson2019jumprelu}, $\sigma_\theta(z)$, illustrated in \cref{fig:jump_relu}. To be precise, using the weight tying scheme of \cref{eq:tying_scheme}, $\tilde{\mathbf{f}}(\mathbf{x})$ can be re-expressed as $\tilde{\mathbf{f}}(\mathbf{x}) = \sigma_{\boldsymbol\theta}(\Wmag\cdot\mathbf{x} + \bmag)$, with the Jump ReLU gap given by $\boldsymbol\theta = \bmag - e^{\rmag} \odot \bgate$; see \cref{app:gated_equiv} for an explanation. We think this is a useful intuition for reasoning about how Gated SAEs reconstruct activations in practice. See Appendix \ref{app:toy-model} for a walkthrough of a toy example where an SAE with Jump ReLUs outperforms one with standard ReLUs.

\subsubsection{Training Gated SAEs}
\label{sec:gated_training}

A natural idea for training gated SAEs would be to apply \cref{eqn:sae_loss}, while restricting the sparsity penalty to just $\mathbf{f}_\text{gate}$:
\begin{equation*}
\mathcal{L}_\text{incorrect}(\mathbf{x}) :=
\underbrace{\norm{\mathbf{x} - \hat{\mathbf{x}}\left(\tilde{\mathbf{f}}(\mathbf{x})\right)}_2^2}_{\mathcal{L}_\text{reconstruct}}
+ \underbrace{\lambda \norm{\fgate(\mathbf{x})
}_1}_{\mathcal{L}_\text{sparsity}}
\end{equation*}
Unfortunately, due to the Heaviside step activation function in $\fgate$, no gradients would propagate to $\Wgate$ and $\bgate$. To mitigate this for the sparsity penalty, we instead apply the L1 norm to the positive parts of the preactivation, $\text{ReLU}\left(\pigate(\mathbf{x})\right)$. To ensure $\fgate$ aids reconstruction by detecting active features, we add an auxiliary task requiring that these same rectified preactivations can be used by the decoder to produce a good reconstruction:
\begin{equation}
\mathcal{L}_\text{gated}(\mathbf{x}) :=
\underbrace{\norm{\mathbf{x} - \hat{\mathbf{x}}\left(\tilde{\mathbf{f}}(\mathbf{x})\right)}_2^2}_{\mathcal{L}_\text{reconstruct}}
+ \underbrace{\lambda \norm{\relu(\pigate(\mathbf{x}))}_1}_{\mathcal{L}_\text{sparsity}}
+ \underbrace{\norm{\mathbf{x} - \hat{\mathbf{x}}_\text{frozen}\left(\relu\left(\pigate(\mathbf{x})\right)\right)}_2^2}_{\mathcal{L}_\text{aux}}
\label{eq:gatedloss}
\end{equation}
where $\hat{\mathbf{x}}_\text{frozen}$ is a frozen copy of the decoder, $\hat{\mathbf{x}}_\text{frozen}(\mathbf{f}) := \Wdec^{\text{copy}} \mathbf{f} + \bdec^{\text{copy}}$, to ensure that gradients from $\mathcal{L}_\text{aux}$ do not propagate back to $\Wdec$ or $\bdec$. This can typically be implemented by stop gradient operations rather than creating copies -- see \cref{app:pseudocode} for pseudo-code for the forward pass and loss function.

To calculate this loss (or its gradient), we have to run the decoder twice: once to perform the main reconstruction for $\mathcal{L}_\text{reconstruct}$ and once to perform the auxiliary reconstruction for $\mathcal{L}_\text{aux}$. This leads to a 50\% increase in the compute required to perform a training update step. However, the increase in overall training time is typically much less, as in our experience much of the training wall clock time goes to generating language model activations (if these are being generated on the fly) or disk I/O (if training on saved activations).

\section{Evaluation}
\label{sec:evaluation}

In this section we benchmark Gated SAEs across a large variety of models and at different sites (\Cref{subsec:benchmarking}), show that they resolve the shrinkage problem (\Cref{subsec:shrinkage}), and show that they produce features that are similarly interpretable to baseline SAE features according to expert human raters, although we could not conclusively determine whether one is better than the other (\Cref{subsec:manual_interp}).

\subsection{Comprehensive Benchmarking}
\label{subsec:benchmarking}

In this subsection we show that Gated SAEs are a Pareto improvement over baseline SAEs on the loss recovered and L0 metrics (\Cref{subsec:background_evaluation}). We show this by evaluating SAEs trained to reconstruct:
\begin{compactenum}
\item The MLP neuron activations in GELU-1L, which is the closest direct comparison to \citet{bricken2023monosemanticity};
\item The MLP outputs, attention layer outputs (taken pre $W_O$ as in \citet{attention_saes}) and residual stream activations in 5 different layers throughout Pythia-2.8B and four different layers in the Gemma-7B base model.
\end{compactenum}
In both experiments, we vary the L1 coefficient $\lambda$ (\Cref{subsec:background_training}) used to train the SAEs, which enables us to compare the Pareto frontiers of L0 and loss recovered between Gated and baseline SAEs.

\begin{figure}[t]
\centering
\includegraphics[width=\textwidth]{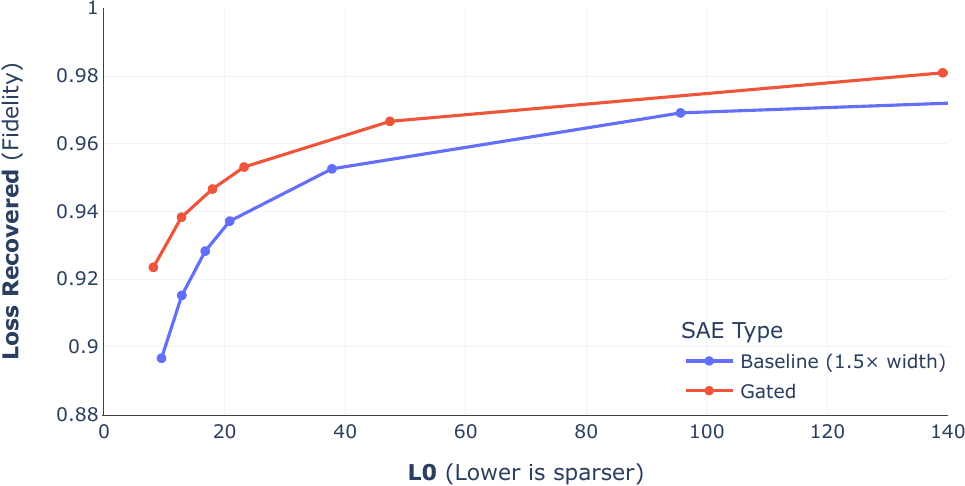}
\caption{Gated SAEs offer better reconstruction fidelity (as measured by loss recovered) at any given level of feature sparsity (as measured by L0). This plot compares Gated and baseline SAEs trained on GELU-1L neuron activations; see \cref{app:more_paretos} for comparisons on Pythia-2.8B and Gemma-7B.}
\label{fig:gelu_1l_eval}
\end{figure}

Gated SAEs require at most 1.5$\times$ more compute to train than regular SAEs (\Cref{sec:gated_training}). To therefore ensure fair comparison in our evaluations, we compare Gated SAEs to baseline SAEs with 50\% more learned features. We show the results for GELU-1L in  \Cref{fig:gelu_1l_eval} and the results for Pythia-2.8B and Gemma-7B in \Cref{app:more_paretos}. In \Cref{app:more_paretos} (\Cref{fig:pythia_full}), at all sites tested, Gated SAEs are a Pareto improvement over regular SAEs. In some cases in \Cref{fig:pythia_full} and \ref{fig:gemma_full} there is a non-monotonic Pareto frontier. We attribute this to difficulties training SAEs (\Cref{app:lessons}).

\subsection{Shrinkage}
\label{subsec:shrinkage}

As described in \cref{subsec:motivation}, the L1 sparsity penalty used to train baseline SAEs causes feature activations to be systematically underestimated, a phenomenon called \emph{shrinkage}. Since this in turn shrinks the reconstructions produced by the SAE decoder, we can observe the extent to which a trained SAE is affected by shrinkage by measuring the average norm of its reconstructions.

Concretely, the metric we use is the \emph{relative reconstruction bias},
\begin{equation}
\gamma := \argmin_{\gamma'} \mathbb{E}_{\mathbf{x}\sim\mathcal{D}}\left[\norm{\hat{\mathbf{x}}_\text{SAE}(\mathbf{x}) / \gamma' - \mathbf{x}}_2^2\right],
\label{eq:relative_norm_bias_defn}
\end{equation}
i.e.~$\gamma^{-1}$ is the optimum multiplicative factor by which an SAE's reconstructions should be rescaled in order to minimise the L2 reconstruction loss; $\gamma=1$ for an unbiased SAE and $\gamma<1$ when there's shrinkage.\footnote{We have defined $\gamma$ this way round so that $\gamma<1$ intuitively corresponds to shrinkage.} Explicitly solving the optimization problem in \cref{eq:relative_norm_bias_defn}, the relative reconstruction bias can be expressed analytically in terms of the mean SAE reconstruction loss, the mean squared norm of input activations and the mean squared norm of SAE reconstructions, making $\gamma$ easy to compute and track during training:\footnote{The second equality makes use of the identity $2\mathbf{a}\cdot\mathbf{b} \equiv \norm{\mathbf{a}}_2^2 + \norm{\mathbf{b}}_2^2 - \norm{\mathbf{a} - \mathbf{b}}_2^2$. Note an unbiased reconstruction ($\gamma=1$) therefore satisfies $\mathbb{E}_{\mathbf{x}\sim\mathcal{D}}\left[\norm{\hat{\mathbf{x}}_\text{SAE}\left(\mathbf{x}\right)}_2^2\right] = \mathbb{E}_{\mathbf{x}\sim\mathcal{D}}\left[\norm{\mathbf{x}}_2^2\right] - \mathbb{E}_{\mathbf{x}\sim\mathcal{D}}\left[\norm{\hat{\mathbf{x}}_\text{SAE}\left(\mathbf{x}\right) - \mathbf{x}}_2^2\right]$; in other words, an unbiased but imperfect SAE (i.e.~one that has non-zero reconstruction loss) must have mean squared reconstruction norm that is strictly \emph{less than} the mean squared norm of its inputs \emph{even without shrinkage}. Shrinkage makes the mean squared reconstruction norm even smaller.}
\begin{equation}
\gamma = \frac{\mathbb{E}_{\mathbf{x}\sim\mathcal{D}}\left[\norm{\hat{\mathbf{x}}_\text{SAE}\left(\mathbf{x}\right)}_2^2\right]}
{\mathbb{E}_{\mathbf{x}\sim\mathcal{D}}\Big[\hat{\mathbf{x}}_\text{SAE}\left(\mathbf{x}\right)\cdot\mathbf{x}\Big]}
= \frac{2\,\mathbb{E}_{\mathbf{x}\sim\mathcal{D}}\left[\norm{\hat{\mathbf{x}}_\text{SAE}\left(\mathbf{x}\right)}_2^2\right]}
{\mathbb{E}_{\mathbf{x}\sim\mathcal{D}}\left[\norm{\hat{\mathbf{x}}_\text{SAE}\left(\mathbf{x}\right)}_2^2\right] + \mathbb{E}_{\mathbf{x}\sim\mathcal{D}}\left[\norm{\mathbf{x}}_2^2\right] - \mathbb{E}_{\mathbf{x}\sim\mathcal{D}}\left[\norm{\hat{\mathbf{x}}_\text{SAE}\left(\mathbf{x}\right) - \mathbf{x}}_2^2\right]}.
\end{equation}

\begin{figure}[t]
\centering
\includegraphics[width=\textwidth]{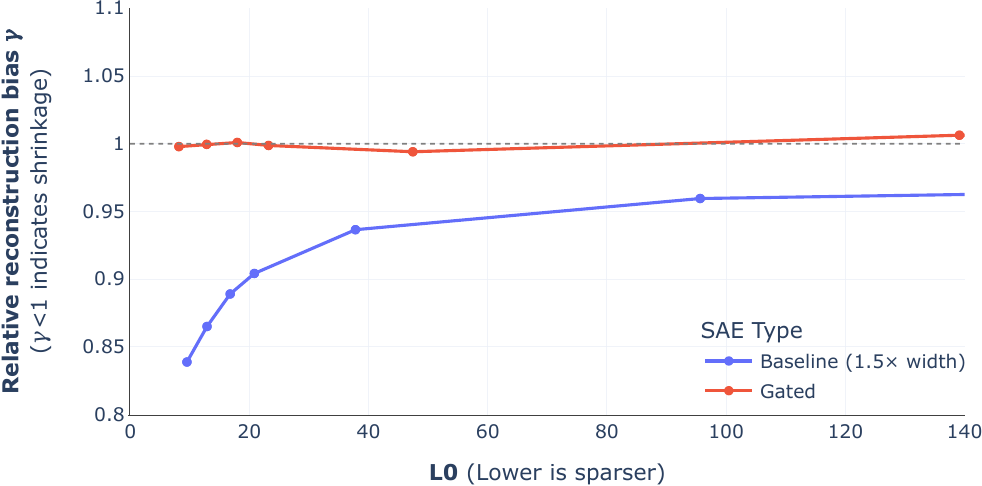}
\caption{Gated SAEs address the shrinkage (GELU-1L neuron activations).}
\label{fig:shrinkage_eval}
\end{figure}

As shown in \Cref{fig:shrinkage_eval}, Gated SAEs' reconstructions are unbiased, with $\gamma\approx1$, whereas baseline SAEs exhibit shrinkage ($\gamma < 1$), with the impact of shrinkage getting worse as the L1 coefficient $\lambda$ increases (and L0 consequently decreases). In \Cref{app:shrinkage_plots} we show that this result generalizes to Pythia-2.8B.

\subsection{Manual Interpretability Scores}
\label{subsec:manual_interp}

\subsubsection{Experimental Methodology}
While we believe that the metrics we have investigated above convey meaningful information about an SAE's quality, they are only imperfect proxies.
As of now, there is no consensus on how to gauge the degree to which a learned feature is `interpretable'. To gain a more qualitative understanding of the difference between the learned dictionary feature, we conduct a blinded human rater experiment, in which we rated the interpretability of a set of randomly sampled features.

We study a variety of SAEs from different layers and sites.

For Pythia-2.8B we had 5 raters, who each rated one feature from baseline and Gated SAEs trained on each (Site, Layer) pair from \Cref{fig:pythia_full}, for a total of 150 features.
For Gemma-7B we had 7 raters; one rated 2 features each, and the rest 1 feature each, from baseline or Gated SAEs trained on each (Site, Layer) pair from \Cref{fig:gemma_full}, for a total of 192 features.

In both cases, the raters were shown the features in random order, without revealing what SAE, site\footnote{Except due to a debugging issue, Gemma attention SAEs were rated separately, so raters were not blind to that.}, or layer they came from. To assess a feature, the rater needed to decide whether there is an explanation of the feature's behavior, in particular for its highest activating examples.
The rater then entered that explanation (if applicable) and selected whether the feature is interpretable (`Yes'), uninterpretable (`No') or maybe interpretable (`Maybe'). As an interface we use an open source SAE visualizer library~\citep{sae_vis}.

\subsubsection{Statistical Analysis}

To test whether Gated SAEs may be more interpretable and estimate the difference, we pair our datapoints according to all covariates (model, layer, site, rater); this lets us control for all of them without making any parametric assumptions, and thus reduces variance in the comparison. We use a one-sided paired Wilcoxon-Pratt signed-rank test, and provide a 90\% BCa bootstrap confidence interval for the mean difference between Baseline and Gated labels, where we count `No' as 0, `Maybe' as 1, and `Yes' as 2. Overall the test of the null hypothesis that Gated SAEs are at most as interpretable as Baseline SAEs gets $p=.060$ (estimate .13, mean difference CI $[0, .26]$). This breaks down into $p=.15$ on just the Pythia-2.8B data (mean difference CI $[-.07, .33]$), and $p=.13$ on just the Gemma-7B data (mean difference CI $[-.04, .29]$).

A Mann-Whitney U rank test on the label differences, comparing results on the two models, fails to reject ($p=.95$) the null hypothesis that they're from the same distribution; the same test directly on the labels similarly fails to reject ($p=.84$) the null hypothesis that they're similarly interpretable overall.

\begin{figure}

\begin{minipage}[b]{0.65\textwidth} 
\centering
\includegraphics[width=\textwidth]{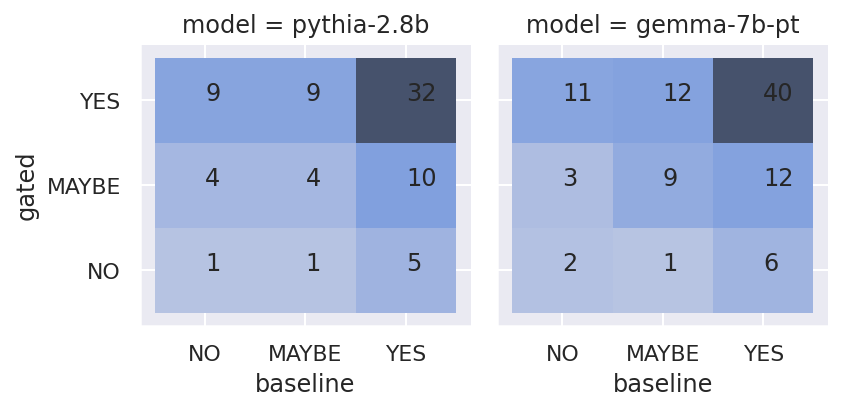}
\end{minipage}
\hfill
\begin{minipage}[b]{0.33\textwidth}
\caption{ \raggedright Contingency table showing Gated vs Baseline interpretability labels from our paired study results, for Pythia-2.8B and Gemma-7B.}
\label{fig:contingency}
\end{minipage}
\end{figure}

The contingency tables used for these results are shown in \Cref{fig:contingency}. The overall conclusion is that, while we can't definitively say the Gated SAE features are more interpretable than those from the Baseline SAEs, they are at least comparable. We provide more analysis of how these break down by site and layer in \Cref{app:study}.

\section{Why do Gated SAEs improve SAE training?}
\label{sec:why_do_gated_sae_improve_sae_training}

In this section we describe an ablation study that reveals the important parts of Gated SAE training (\Cref{subsec:ablation_study}) and benchmark Gated SAEs against a closely related approach to resolving shrinkage (\Cref{subsec:rescale_expt}).

\subsection{Ablation Study}
\label{subsec:ablation_study}

\begin{figure}[t]
\centering
\includegraphics[width=\textwidth]{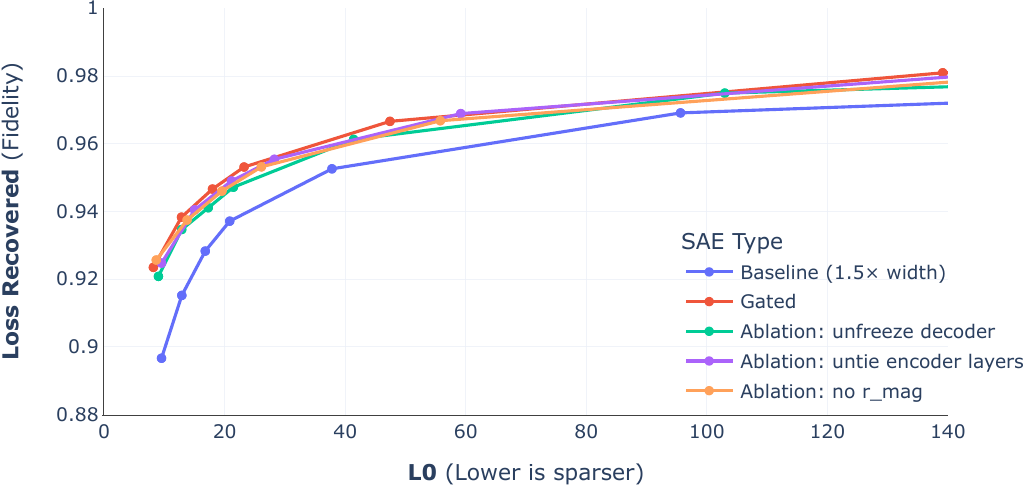}
\caption{Our ablation study on GELU-1L MLP neuron activations indicates: (a) the importance of freezing the decoder in the auxiliary task $\mathcal{L}_\text{aux}$ used to train $\mathbf{f}_\text{gate}$'s parameters; (b) tying encoder weights according to \cref{eq:tying_scheme} is slightly beneficial for performance (in addition to yielding a significant reduction in parameter count and inference compute); (c) further simplifying the encoder weight tying scheme in \cref{eq:tying_scheme} by removing $\rmag$ is mildly harmful to performance.}
\label{fig:ablation_study}
\end{figure}

In this section, we vary several parts of the Gated SAE training methodology (\Cref{subsec:defining_gated_saes}) to gain insight into which aspects of the training are required for the observed improvement in performance. Gated SAEs differ from baseline SAEs in many respect, making it easy to incorrectly attribute the performance gains to spurious details without a careful ablation study. \Cref{fig:ablation_study} shows Pareto frontiers for these variations and below which we describe each variation in turn and discuss our interpretation of the results.

\begin{compactenum}
\item \textbf{Unfreeze decoder}: Here we unfreeze the decoder weights in $\mathcal{L}_\text{aux}$ -- i.e.~allow this auxiliary task to update the decoder weights in addition to training $\mathbf{f}_\text{gate}$'s parameters. Although this (slightly) simplifies the loss, there is a reduction in performance, providing evidence in support of the hypothesis that it is beneficial to limit the impact of the L1 sparsity penalty to just those parameters in the SAE that need it -- i.e.~those used to detect which features are active.
\item \textbf{No $\rmag$}: Here we remove the $\rmag$ scaling parameter in \cref{eq:tying_scheme}, effectively setting it to zero (so that we multiply by $e^0=1$); this further ties $\mathbf{f}_\text{gate}$'s and $\mathbf{f}_\text{mag}$'s parameters together. With this change, the two encoder sublayers' preactivations can at most differ by an elementwise shift.\footnote{Because the two biases $\bgate$ and $\bmag$ can still differ.} There is a slight drop in performance, suggesting $\rmag$ contributes somewhat (but not critically) to the improved performance of the Gated SAE.
\item \textbf{Untied encoders}: Here we check whether our choice to share the majority of parameters between the two encoders has meaningfully hurt performance, by training Gated SAEs with gating and ReLU encoder parameters completely untied. Despite the greater expressive power of an untied encoder, we see no improvement in performance -- in fact a slight deterioration. This suggests our tying scheme (\cref{eq:tying_scheme}) -- where encoder directions are shared, but magnitudes and biases aren't -- is effective at capturing the advantages of using a gated SAE while avoiding the 50\% increase in parameter count and inference-time compute of using an untied SAE.
\end{compactenum}

\subsection{Is it sufficient to just address shrinkage?}
\label{subsec:rescale_expt}

As explained in \cref{subsec:motivation}, SAEs trained with the baseline architecture and L1 loss systematically underestimate the magnitudes of latent features' activations (i.e.~shrinkage). Gated SAEs, through modifications to their architecture and loss function, overcome these limitations, thereby addressing shrinkage.

It is natural to ask to what extent the performance improvement of Gated SAEs is solely attributable to addressing shrinkage. Although addressing shrinkage would -- all else staying equal -- improve reconstruction fidelity, it is not the only way to improve SAEs' performance: for example, gated SAEs could also improve upon baseline SAEs by learning better encoder directions (for estimating when features are active and their magnitudes) or by learning better decoder directions (i.e.~better dictionaries for reconstructing activations).

In this section, we try to answer this question by comparing Gated SAEs trained as described in \cref{sec:gated_training} with an alternative (architecturally equivalent) approach that also addresses shrinkage, but in a way that uses frozen encoder and decoder directions from a baseline SAE of equal dictionary size.\footnote{Concretely, we do this by training baseline SAEs, freezing their weights, and then learning additional rescale and shift parameters (similar to \citet{wright2024addressing}) to be applied to the (frozen) encoder pre-activations before estimating feature magnitudes.} Any performance improvement over baseline SAEs obtained by this alternative approach (which we dub ``baseline + rescale \& shift'') can only be due to better estimations of active feature magnitudes, since by construction an SAE parameterised by ``baseline + rescale \& shift'' shares the same encoder and decoder directions as a baseline SAE.

\begin{figure}[t]
\centering
\includegraphics[width=\textwidth]{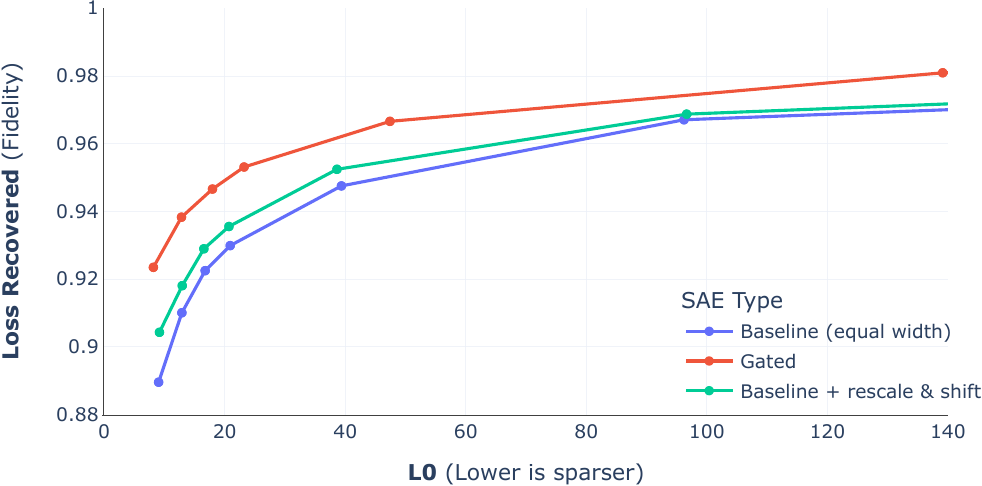}
\caption{Evidence from GELU-1L that the performance improvement of gated SAEs does not solely arise from addressing shrinkage (systematic underestimation of latent feature activations). Taking a frozen baseline SAE's parameters and learning $\rmag$ and $\bmag$ parameters on top of them (green line) does successfully resolve shrinkage, by decoupling feature magnitude estimation from active feature detection. However, it explains only a small part of the performance increase of gated SAEs (red line) over baseline SAEs (blue line).}
\label{fig:rescale_shift_ablation}
\end{figure}

As shown in \cref{fig:rescale_shift_ablation}, although resolving shrinkage only (``baseline + rescale \& shift'') does improvement baseline SAEs' performance a little, a significant gap remains with respect to the performance of gated SAEs. This suggests that the benefit of the gated architecture and loss comes from learning better encoder and decoder directions, not just from overcoming shrinkage. In \cref{app:inference_time_optimization} we explore further how Gated and baseline SAEs' decoders differ by replacing their respective encoders with an optimization algorithm at inference time.
\section{Related Work}
\label{sec:related_work}

\textbf{Mechanistic Interpretability}. We hope that our improvements to Sparse Autoencoders are helpful for mechanistic interpretability research. Recent mechanistic interpretability work has found recurring components in small and large LMs \citep{olsson2022context}, identified computational subgraphs that carry out specific tasks in small LMs (circuits; \citet{wang2023interpretability}) and reverse-engineered how toy tasks are carried out in small transformers \citep{nanda2023progress}. Limitations of existing work include (i) how they only study narrow subsets of the natural language training distribution are studied (though see \citet{copy_suppression}) and (ii) current work has not explained how frontier language models function mechanistically \citep{gpt4, gemini, claude3}. SAEs may be key to explaining model behaviour across the whole training distribution \citep{bricken2023monosemanticity} and are trained without supervision, which may enable future work to explain how larger models function on broader tasks.

\textbf{Classical Dictionary Learning}. Our work builds on a large amount of research that precedes transformers, and even deep learning. For example, sparse coding \citep{elad2010sparse} studies how discrete and continuous representations can involve more representations than basis vectors, like our setup in \Cref{sec:intro}, and sparse representations are also studied in neuroscience \citep{thorpe1989local, olshausen}. Further, shrinkage (\Cref{subsec:shrinkage}) is built into the Lasso \citep{tibshirani1996regression} and well-studied in statistical learning \citep{sparsitybook}. One dictionary learning algorithm, k-SVD \citep{aharon2006k} also uses two stages to learn a dictionary like Gated SAEs.

\textbf{Dictionary Learning in Language Models}. Early work into applying Dictionary Learning to LMs include \citet{sharkey2022interim} (on a GPT-2-like model), \citet{yun2023transformer} (on a BERT model), \citet{tamkin2023codebook} (with discrete features, and during LM pretraining) and \citet{cunningham2023sparse} (on a small Pythia model). \citet{bricken2023monosemanticity}'s work later provided a widely-scoped analysis of SAEs trained on a 1L model, evaluating the loss when splicing the SAE into the forward pass (\Cref{subsec:benchmarking}), evaluating the impact of learned features on LM rollouts, and visualizing and interpreting all learned features with autointerpretability \citep{bills2023language}. Following this work, other researchers have extended SAE training to attention layer outputs \citep{attention_saes, gpt2_attention_saes} and residual stream states \citep{bloom2024gpt2residualsaes}.

\textbf{Dictionary Learning's Limitations and Improvements}. \citet{wright2024addressing} raised awareness of shrinkage (\Cref{subsec:shrinkage}) and proposed addressing this via decoder finetuning. A difficulty with this approach is that it is not possible to fine tune \emph{all} the SAEs parameters in this way without losing sparsity and/or interpretability of feature directions. This limits the extent to which fine-tuning can remove the biases baked into the SAEs parameters during L1-based pre-training. Gated SAEs address this issue (\Cref{subsec:defining_gated_saes}). \citet{marks2024sparse} stress-test how useful SAEs are, and find success but also rely on methods that leave many error nodes in their computational subgraphs, which represent the difference between SAE reconstructions and the ground truth. A series of updates to the work in \citet{bricken2023monosemanticity} have also proposed SAE training methodology improvements \citep{anthropic_january_update, anthropic_february_update, anthropic_march_update}. In parallel to our work, \citet{taggart} finds early improvements with a similar Jump ReLU \citep{erichson2019jumprelu} architecture change to SAEs, but with a different loss function, and without addressing the problems of L1.

\textbf{Disentanglement} \citep{bengio2013deep} aims to learn representations that separate out distinct, independent `factors of variation' of the underlying data generating process.
This is somewhat similar to our aims with dictionary learning, as we want to separate an activation vector into distinct, sparse factors of variation (weights on feature directions), although the dictionary elements are not completely independent, as it may not be possible to accurately represent two features simultaneously due to interference between non-orthogonal dictionary features.
Methods explicitly motivated by learning a disentangled representation typically enforce a prior structure on the learned representation, typically that features are aligned with the basis of a latent space \citep{kim2018disentangling, mathieu2019disentangling,chen2016infogan,chen2018isolating}.
In contrast, in our work we focus on the representation space of a pre-trained language model, rather than trying to learn a representation directly from data, and enforce a different prior structure, of decomposition into a sparse linear combination of an overcomplete basis.
In a sense, our work proceeds from the theory that language models have succeeded in learning a disentangled representation of the data with a particular structure, which we are trying to recover.

\section{Conclusion}
\label{sec:conclusion}

In this work we introduced Gated SAEs (\Cref{subsec:defining_gated_saes}) which are a Pareto improvement in terms of reconstruction quality and sparisty compared to baseline SAEs (\Cref{subsec:benchmarking}), and are comparably interpretable (\Cref{subsec:manual_interp}). We showed via an ablation study that every key part of the Gated SAE methodology was necessary for strong performance (\Cref{subsec:ablation_study}). This represents significant progress on improving Dictionary Learning in LLMs -- at many sites, Gated SAEs require half the L0 to achieve the same loss recovered (\Cref{fig:pythia_full}). This is likely to improve work that uses SAEs to steer language models \citep{gdm_progress_update_1}, interpret circuits \citep{marks2024sparse}, or understand language model components across the full distribution \citep{bricken2023monosemanticity}.

\textbf{Limitations}. Our work, like all sparse autoencoder research, is motivated by several assumptions about the sparsity and linearity of computation in Large Language Models (\Cref{sec:intro}). If these assumptions are false, our work may still be useful (see footnote 1), but we may be making incorrect conclusions from work using SAEs, since they bake in the sparsity and linearity assumptions. Separately, our work complicates SAE training with a more complex encoder.

One worry about increasing the expressivity of sparse autoencoders is that they will overfit when reconstructing activations \citep[Dictionary Learning Worries]{anthropic_may_update}, since the underlying model only uses simple MLPs and attention heads, and in particular lacks discontinuities such as step functions. Overall we do not see evidence for this. Our evaluations use held-out test data and we check for interpretability manually. But these evaluations are not totally comprehensive: for example, they do not test that the dictionaries learned contain causally meaningful intermediate variables in the model's computation. The discontinuity in particular introduces issues with methods like integrated gradients \citep{sundararajan2017axiomatic} that discretely approximate a path integral, as applied to SAEs by \citet{marks2024sparse}.

Finally, it could be argued that some of the performance gap between Gated and baseline SAEs could be closed by inexpensive inference-time interventions that prune the many low activating features that tend to appear in baseline SAEs -- because baseline SAEs don't have a thresholding mechanism like Gated SAEs do (\cref{app:gated_equiv}). Without such interventions, these low activating features increase baseline SAEs' L0 at a given loss recovered without contributing much to reconstruction (due to low magnitude), and with unclear impact on interpretability.

\textbf{Future work}. Future work could verify that Gated SAEs continue to improve dictionary learning beyond 7B base LLMs, such as by extending to larger chat models, or even to multimodal or Mixture-of-Experts models. Alternatively, work could look into the features learned by Gated and baseline SAEs and determine whether the architectures have differences in inductive biases beyond those we noted in this work. We expect it may be possible to further improve Gated SAEs' performance through additional tweaks to the architecture and training procedure. Finally, we would be most excited to work on using dictionary learning techniques to further interpretability in general, such as to improve circuit finding \citep{conmy2023automated,marks2024sparse} or steering \citep{actadd} in language models, and hope that Gated SAEs can serve to accelerate such work.

\section{Acknowledgements}

We would like to thank Romeo Valentin for conversations that got us thinking about k-SVD in the context of SAEs, which inspired part of our work. Additionally, we are grateful for Vladimir Mikulik's detailed feedback on a draft of this work which greatly improved our presentation, and Nicholas Sonnerat's work on our codebase and help with feature labelling. We would also like to thank Glen Taggart who found in parallel work \citep{taggart} that a similar method gave improvements to SAE training, helping give us more confidence in our results. Finally, we are grateful to Sam Marks for pointing out an error in the derivation of relative reconstruction bias in an earlier version of this paper.

\section{Author contributions}

Senthooran Rajamanoharan developed the Gated SAE architecture and training methodology, inspired by discussions with Lewis Smith on the topic of shrinkage. Arthur Conmy and Senthooran Rajamanoharan performed the mainline experiments in \cref{sec:evaluation} and \cref{sec:why_do_gated_sae_improve_sae_training} and led the writing of all sections of the paper. Tom Lieberum implemented the manual interpretability study of \cref{subsec:manual_interp}, which was designed and analysed by J\'anos Kram\'ar. Tom Lieberum also created \cref{fig:gated_sae_architecture}. Lewis Smith contributed \cref{app:inference_time_optimization} and Neel Nanda contributed \cref{app:toy-model}. Our SAE codebase was designed by Vikrant Varma who implemented it with Tom Lieberum, and was scaled to Gemma by Arthur Conmy, with contributions from Senthooran Rajamanoharan and Lewis Smith. J\'anos Kram\'ar built most of our underlying interpretability infrastructure. Rohin Shah and Neel Nanda edited the manuscript and provided leadership and advice throughout the project.

\bibliography{main}
\section*{Appendix}
\appendix

\section{Inference-time optimization}
\label{app:inference_time_optimization}

The task SAEs perform can be split into two sub-tasks: sparse coding, or learning a set of features from a dataset, and sparse approximation, where a given datapoint is approximated as a sparse linear combination of these features.
The decoder weights are the set of learned features, and the mapping represented by the encoder is a sparse approximation algorithm.
Formally, sparse approximation is the problem of finding a vector $\boldsymbol\alpha$ that minimises;
\begin{align}
\boldsymbol\alpha = \arg \min \norm{\mathbf{x} - \mathbf{D} \boldsymbol\alpha}^2_2\;\;s.t.\;\;\norm{\boldsymbol\alpha}_0 < \gamma
\label{eqn:normal_sparse_approx}
\end{align}
i.e.~that best reconstructs the signal $\mathbf{x}$ as a linear combination of vectors in a dictionary $\mathbf{D}$, subject to a constraint on the L0 pseudo-norm on $\boldsymbol\alpha$. 
Sparse approximation is a well studied problem, and SAEs are a \textit{weak} sparse approximation algorithm.
SAEs, at least in the formulation conventional in dictionary learning for language models, in fact solve a slightly more restricted version of this problem where the weights $\boldsymbol\alpha$ on each feature are constrained to be non-negative, leading to the related problem 
\begin{align}
\boldsymbol\alpha = \arg \min \norm{\mathbf{x} - \mathbf{D} \boldsymbol\alpha}^2_2 \;\;s.t.\;\;\norm{\boldsymbol\alpha}_0 < \gamma, \boldsymbol{\alpha} > 0
\label{eqn:sae_sparse_approx}
\end{align}

In this paper, we do not explore using more powerful algorithms for sparse coding.
This is partly because we are using SAEs not just to recover \textit{a} sparse reconstruction of activations of a LM; ideally we hope that the learned features will coincide with the linear representations actually used by the LM, under the superposition hypothesis.
Prior work \citep{bricken2023monosemanticity} has argued that SAEs are more likely to recover these due to the correspondence between the SAE encoder and the structure of the network itself; the argument is that it is implausible that the network can make use of features which can only be recovered from the vector via an iterative optimisation algorithm, whereas the structure of the SAE means that it can only find features whose presence can be predicted well by a simple linear mapping.
Whether this is true remains, in our view, an important question for future work, but we do not address it in this paper.

In this section we discuss some results obtained by using the dictionaries learned via SAE training, but replacing the encoder with a different sparse approximation algorithm at inference time.
This allows us to compare the dictionaries learned by different SAE training regimes independently of the quality of the encoder.
It also allows us to examine the gap between the sparse reconstruction performed by the encoder against the baseline of a more powerful sparse approximation algorithm.
As mentioned, for a fair comparison to the task the encoder is trained for, it is important to solve the sparse approximation problem of \cref{eqn:sae_sparse_approx}, rather than the more conventional formulation of \cref{eqn:normal_sparse_approx}, but most sparse approximation algorithms can be modified to solve this with relatively minor changes.

Solving \cref{eqn:sae_sparse_approx} exactly is equivalent to integer linear programming, and is NP hard.
The integer linear programs in question would be large, as our SAE decoders routinely have hundreds of thousands of features, and solving them to guaranteed optimality would likely be intractable.
Instead, as is commonly done, we use iterative greedy algorithms to find an approximate solution.
While the solution found by these sparse approximation algorithms is not guaranteed to be the global optimum, these are significantly more powerful than the SAE encoder, and we feel it is acceptable in practice to treat them as an upper bound on possible encoder performance.

For all results in this section, we use gradient pursuit, as described in \citet{blumensath2008gradient}, as our inference time optimisation (ITO) algorithm.
This algorithm is a variant of orthogonal matching pursuit \citep{pati1993orthogonal} which solves the orgothonalisation of the residual to the span of chosen dictionary elements approximately at every step rather than exactly, but which only requires matrix multiplies rather than matrix solves and is easier to implement on accelerators as a result.
It is possibly not crucial for performance that our optimisation algorithm be implementable on TPUs, but being able to avoid a host-device transfer when splicing this into the forward pass allowed us to re-use our existing evaluation pipeline with minimal changes.

\begin{figure}
    \centering
    \includegraphics[width=1.0\textwidth]{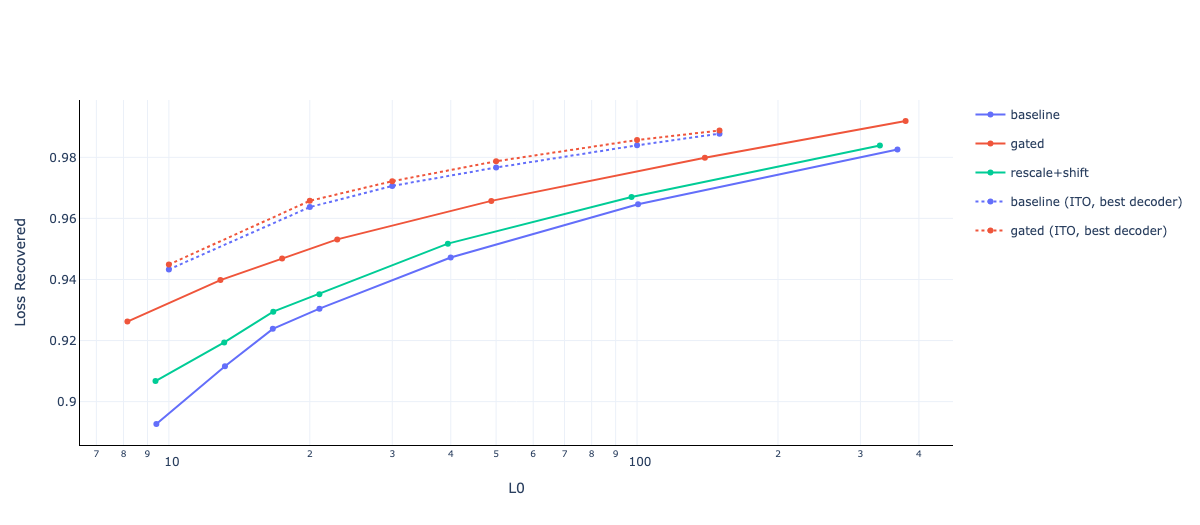}
    \caption{Pareto frontiers of a baseline SAE, a baseline SAE with learned rescale and shift (to account for shrinkage) and 
    a gated SAE across different sparsity lambdas, compared to the ITO Pareto frontier of the best decoder of each type with ITO, varying the target sparsity. The best gated encoder is better than the best standard encoder by this measure, but the difference is marginal. As shown in the plot above, the best baseline encoder by the ITO measure had a much larger test time sparsity (around 100) than the best gated model (around 30).
    This figure suggests that the gap between SAE performance and 'optimal' performance, if we assume that ITO is close to the maximum possible reconstruction using the given encoder, is much smaller for the gated model.
    }
    \label{fig:pareto_ito}
\end{figure}

When we use a sparse approximation algorithm at test time, we simply use the decoder of a trained SAE as a dictionary, ignoring the encoder.
This allows us to sweep the target sparsity at test time without retraining the model, meaning that we can plot an entire Pareto frontier of loss recovered against sparsity for a single decoder, as in done in \Cref{fig:pareto_ito}.

\begin{figure}
    \centering
    \includegraphics[width=0.9\textwidth]{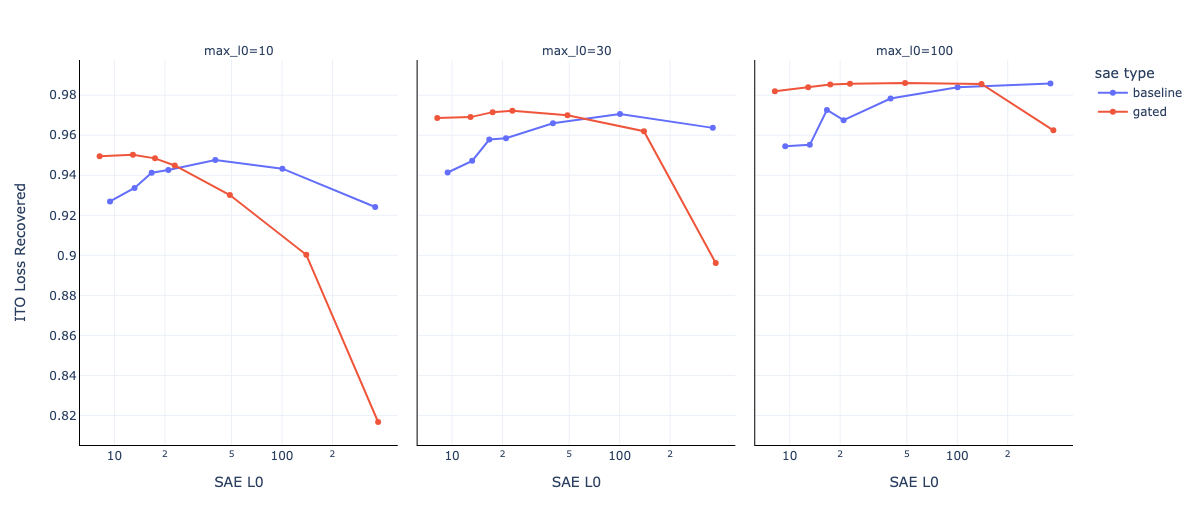}
    \caption{This figure compares the ITO performance of different decoders across a sweep for decoders trained using a baseline SAE and the gated method, at three different test time target sparsities. Gated SAEs trained at lower target sparsities consistently achieve better dictionaries by this measure. Interestingly, the best performing baseline dictionary by this measure often has a much higher test time sparsity than the target; for instance, at a test time sparsity of 30, the best baseline SAE was the one that had a test time sparsity of more like 100.
    This could be an artifact of the fact that the L0 measure is quite sensitive to noise, and standard SAE architectures tend to have a reasonable number of features with very low activation.}
    \label{fig:ito_comparison}
\end{figure}

\Cref{fig:ito_comparison} compares the loss recovered when using ITO for a suite of SAEs decoders trained with both methods at three different test time L0 thresholds.
This graph shows a somewhat surprising result; while Gated SAEs learn better decoders generally, and often achieve the best loss recovered using ITO close to their training sparsity, SAE decoders are often outperformed by decoders which achieved a higher test time L0; it's better to do ITO with a target L0 of 10 with an decoder with an achieved L0 of around 100 during training than one which was actually trained with this level of sparsity.
For instance, the left hand panel in \Cref{fig:ito_comparison} shows that SAEs with a training L0 of 100 are better than those with an L0 of around 10 at almost every sparsity level in terms of ITO reconstruction. 
However, gated SAE dictionaries have a small but real advantage over standard SAEs in terms of loss recovered at most target sparsity levels, suggesting that part of the advantage of gated SAEs is that they learn better dictionaries as well as addressing issues with shrinkage.
However, there are some subtleties here; for example, we find that baseline SAEs trained with a lower sparsity penalty (higher training L0) often outperform more sparse baseline SAEs according to this measure, and the best performing baseline SAE (L0 $\approx 99$) is comparable to the best performing Gated SAE (L0 $\approx 20$).

\Cref{fig:pareto_ito} compares the Pareto frontiers of a baseline model and a gated model to the Pareto frontier of an ITO sweep of the best performing dictionary of each.
Note that, while the Pareto curve of the baseline dictionary is formed by several models as each encoder is specialised to a given sparsity level, as mentioned, ITO lets us plot a Pareto frontier by sweeping the target sparsity with a single dictionary; here we plot only the best performing dictionary from each model type to avoid cluttering the figure.
This figure suggests that the performance gap between the encoder and using ITO is smaller for the gated model. Interestingly, this cannot solely be explained by addressing shrinkage, as we demonstrate by experimenting with a baseline model which learns a rescale and shift with a frozen encoder and decoder directions. 

\section{More Loss Recovered / L0 Pareto frontiers}
\label{app:more_paretos}

\begin{figure}[hp]
    \centering
    \includegraphics[width=1.05\textwidth]{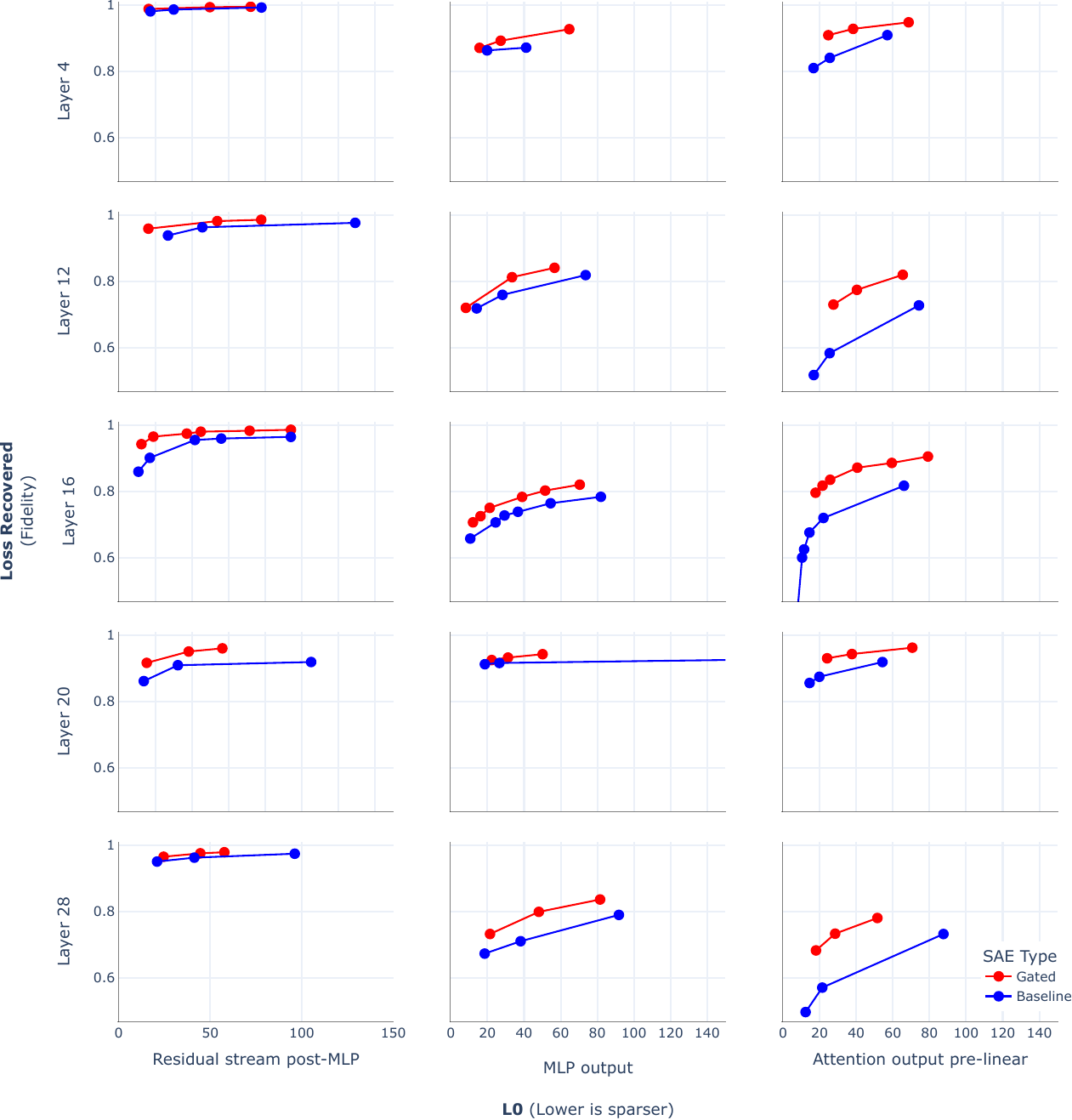}
    \caption{Gated SAEs throughout Pythia-2.8B. At all sites we tested, Gated SAEs are a Pareto improvement. In every plot, the SAE with maximal loss recovered was a Gated SAE.}
    \label{fig:pythia_full}
\end{figure}

\begin{figure}[hp]
    \centering
    \includegraphics[width=1.05\textwidth]{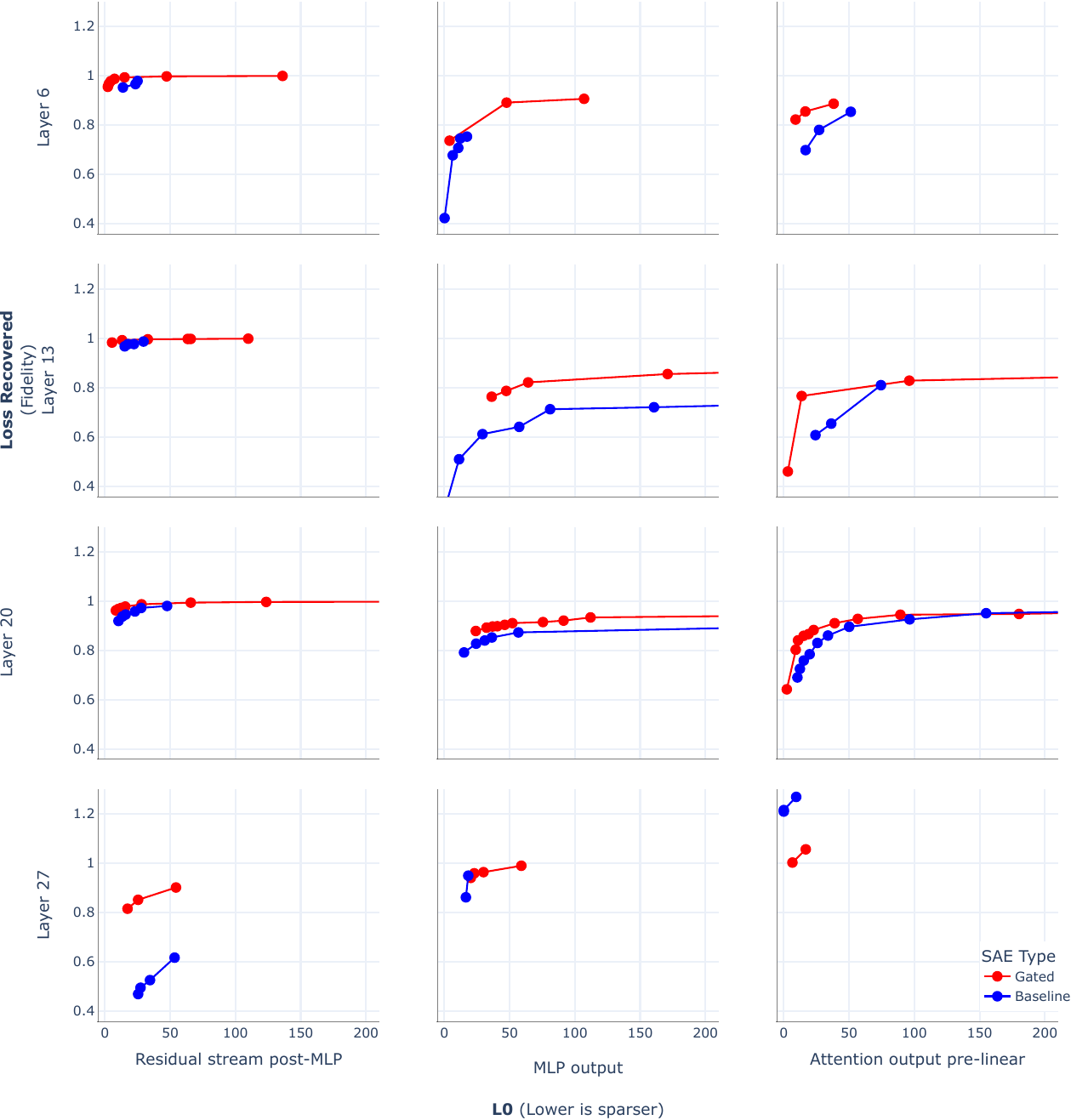}
    \caption{Gated and Normal Pareto-Optimal SAEs for Gemma-7B -- see \Cref{app:more_paretos} for a discussion of the anomalies (such as the Layer 27 attention output SAEs), and \Cref{table:gemma_7b_baselines}-\ref{table:gemma_7b_gated2} for full stats (including points not on the Pareto frontier).}
    \label{fig:gemma_full}
\end{figure}

In \Cref{fig:pythia_full} we show that Gated SAEs outperform baseline SAEs. In \Cref{fig:gemma_full} we show that Gated SAEs ourperform baseline SAEs at all but one MLP output or residual stream site that we tested on.

In \Cref{fig:gemma_full} at the attention output pre-linear site at layer 27, loss recovered is bigger than 1.0. On investigation, we found that the dataset used to train the SAE was not identical to Gemma's pretraining dataset, and at this site it was possible to mean ablate this quantity and decrease loss -- explaining why SAE reconstructions had lower loss than the original model.

\section{Further Shrinkage Plots}
\label{app:shrinkage_plots}

\begin{figure}[hp]
    \centering
    \includegraphics[width=1.05\textwidth]{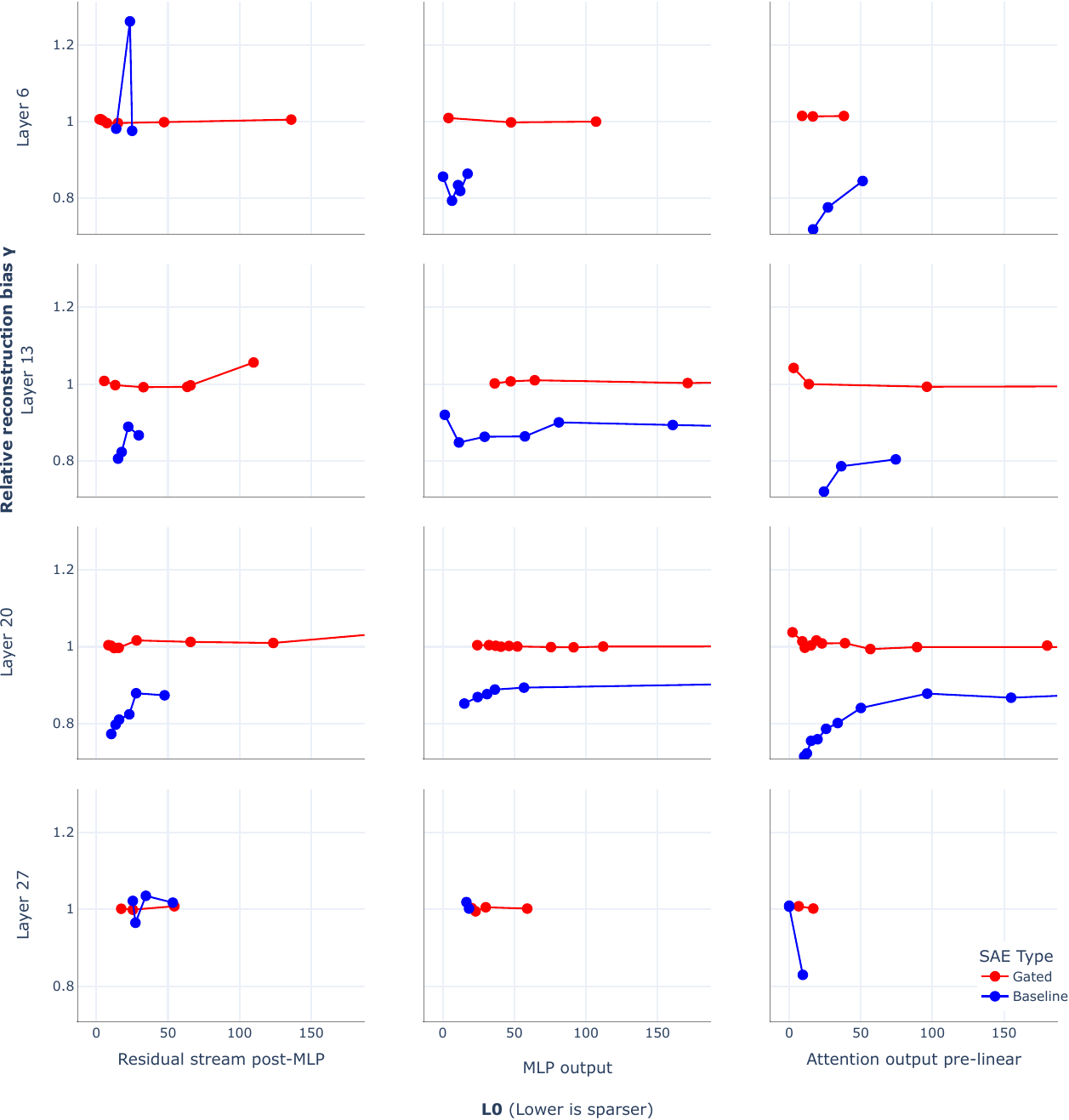}
    \caption{Gated SAEs address the shrinkage problem (\Cref{subsec:shrinkage}) in Pythia-2.8B.}
    \label{fig:pythia_shrinkage}
\end{figure}

In \Cref{fig:pythia_shrinkage}, we show that Gated SAEs resolve shrinkage (as measured by relative reconstruction bias (\Cref{subsec:shrinkage})) in Pythia-2.8B.

\section{Training and evaluation: hyperparameters and other details.}
\label{app:training_hyperparameters_and_details}

\subsection{Training}

\subsubsection{General training details}

Other details of SAE training are:
\begin{compactitem}

\item \textbf{SAE Widths}. Our SAEs have width $2^{17}$ for most baseline SAEs, $3 \times 2^{16}$ for Gated SAEs, except for the (Pythia-2.8B, Residual Stream) sites we used $2^{15}$ for baseline and $3\times 2^{14}$ for Gated since early runs at these sites had lots of learned feature death.

\item \textbf{Training data}. We use activations from hundreds of millions to billions of activations from LM forward passes as input data to the SAE. Following \citet{neel_sae_replication}, we use a shuffled buffer of these activations, so that optimization steps don't use data from highly correlated activations.\footnote{In contrast to earlier findings \citep{arthur_sae_replication}, we found that when using Pythia-2.8B's activations from sequences of length 2048, rather than GELU-1L's activations from sequences of length 128, it was important to shuffle the $10^6$ length activation buffer used to train our SAEs.} 

\item \textbf{Resampling}. We used \textit{resampling}, a technique which at a high-level reinitializes features that activate extremely rarely on SAE inputs periodically throughout training. We mostly follow the approach described in the `Neuron Resampling' appendix of \citet{bricken2023monosemanticity}, except we reapply learning rate warm-up after each resampling event, reducing learning rate to 0.1x the ordinary value, and, increasing it with a cosine schedule back to the ordinary value over the next 1000 training steps.

\item \textbf{Optimizer hyperparameters}. We use the Adam optimizer with $\beta_2 = 0.999$ and $\beta_1 = 0.0$, following \citet{anthropic_february_update}, as we also find this to be a slight improvement to training. We use a learning rate warm-up. See \Cref{subsubapp1} for learning rates of different experiment. 

\item \textbf{Decoder weight norm constraints}. \citet{anthropic_february_update} suggest constraining columns to have \emph{at most} unit norm (instead of exactly unit norm), which can help distinguish between productive and unproductive feature directions (although it should have no systematic impact on performance). However, we follow the original approach of constraining columns to have exact unit norms in this work for the sake of simplicity.

\item \textbf{Interpreting the L1 $\lambda$ coefficients.}. In our infrastructure we calculate L2 loss and then divide by $n$. In the baseline experiments we further divide the reconstruction L2 loss by $\mathbb{E} ||x||_2$.

\end{compactitem}

\subsubsection{Experiment-specific training details}
\label{subsubapp1}

\begin{compactitem}
    
\item We use learning rate 0.0003 for all Gated SAE experiments, and the GELU-1L baseline experiment. We swept for optimal baseline learning rates for the GELU-1L baseline to generate this value. For the Pythia-2.8B and Gemma-7B baseline SAE experiments, we divided the L2 loss by $\mathbb{E} ||x||_2$, motivated by better hyperparameter transfer, and so changed learning rate to 0.001 and 0.00075 (full learning rate detail in tables \Cref{table:gemma_7b_baselines}-\ref{table:pythia_2p8b_gated2}). We didn't see noticeable difference in the Pareto frontier and so did not sweep this hyperparameter further. 

\item We generate activations from sequences of length 128 for GELU-1L, 2048 for Pythia-2.8B and 1024 for Gemma-7B.

\item We use a batch size of 4096 for all runs. We use 300,000 training steps for GELU-1L and Gemma-7B runs, and 400,000 steps for Pythia-2.8B runs. 

\end{compactitem}

\subsubsection{Lessons learned scaling SAEs}
\label{app:lessons}

\begin{compactitem}

\item \textbf{Learned feature death is unpredictable}. In \Cref{table:gemma_7b_baselines} (and other tables) there are few patterns that can be gleaned from staring at which runs have high numbers of dead learned features (called dead neurons in \citet{bricken2023monosemanticity}).

\item \textbf{Resampling makes hyperparameter sweeps difficult}. We found that resampling caused L0 and loss recovered to increase, similar to \citet{arthur_sae_replication}.

\item \textbf{Training appears to converge earlier than expected}. We found that we did not need 20B tokens as in \citet{bricken2023monosemanticity}, as generally resampling had stopped causing gains and loss curves plateaued after just over one billion tokens.


\end{compactitem}

\subsection{Evaluation}
\label{subapp:evaluation}

We evaluated the models on over a million held-out tokens. Tables \ref{table:gemma_7b_baselines}-\ref{table:pythia_2p8b_gated2} show summary stats from training runs on the Pareto frontier.

\begin{table}[ht!]
\centering
\scriptsize
\begin{tabular}{|c|c|c|c|c|c|c|c|c|c|c|c|}
\hline
\textbf{\tiny{Site}} & \textbf{\tiny{Layer}} & \textbf{\tiny{Sparsity $\lambda$}} & \textbf{\tiny{LR}} & \textbf{\tiny{L0}} & \textbf{\tiny{\% CE Recovered}} & \textbf{\tiny{Clean CE Loss}} & \textbf{\tiny{SAE CE Loss}} & \textbf{\tiny{0 Abl. CE Loss}} & \textbf{\tiny{Width}} & \textbf{\tiny{\% Alive Features}} & \textbf{\tiny{Shrinkage $\gamma$}} \\
\hline
\textit{Resid} & \textit{6} & \textit{3e-05} & \textit{0.001} & \textit{18.1} & \textit{95.28\%} & \textit{2.5426} & \textit{3.1847} & \textit{16.1549} & \textit{196608} & \textit{16.8\%} & \textit{0.982} \\
Resid & 6 & 2e-05 & 0.001 & 10.5 & 85.3\% & 2.5426 & 4.5433 & 16.1549 & 196608 & 5.72\% & 1.136 \\
Resid & 6 & 1e-05 & 0.001 & 19.0 & 91.24\% & 2.5426 & 3.7349 & 16.1549 & 196608 & 5.11\% & 1.606 \\
\textit{Resid} & \textit{6} & \textit{2e-05} & \textit{0.00075} & \textit{29.8} & \textit{96.65\%} & \textit{2.5426} & \textit{2.9989} & \textit{16.1549} & \textit{196608} & \textit{13.67\%} & \textit{1.261} \\
\textit{Resid} & \textit{6} & \textit{3e-05} & \textit{0.00075} & \textit{25.4} & \textit{97.9\%} & \textit{2.5426} & \textit{2.8279} & \textit{16.1549} & \textit{196608} & \textit{38.86\%} & \textit{0.976} \\
Resid & 6 & 8e-06 & 0.00075 & 29.8 & 91.28\% & 2.5426 & 3.7301 & 16.1549 & 196608 & 9.88\% & 1.105 \\
Resid & 6 & 1e-05 & 0.00075 & 57.3 & 97.36\% & 2.5426 & 2.9023 & 16.1549 & 196608 & 11.78\% & 1.03 \\
\hline
\end{tabular}
\caption{Gemma-7B Baseline SAEs (1024 sequence length). Italic are Pareto optimal SAEs.}
\label{table:gemma_7b_baselines}
\end{table}

\begin{table}[ht!]
\centering
\scriptsize\begin{tabular}{|c|c|c|c|c|c|c|c|c|c|c|c|}
\hline
\textbf{\tiny{Site}} & \textbf{\tiny{Layer}} & \textbf{\tiny{Sparsity $\lambda$}} & \textbf{\tiny{LR}} & \textbf{\tiny{L0}} & \textbf{\tiny{\% CE Recovered}} & \textbf{\tiny{Clean CE Loss}} & \textbf{\tiny{SAE CE Loss}} & \textbf{\tiny{0 Abl. CE Loss}} & \textbf{\tiny{Width}} & \textbf{\tiny{\% Alive Features}} & \textbf{\tiny{Shrinkage $\gamma$}} \\
\hline
MLP & 20 & 5e-05 & 0.00075 & 26.0 & 82.36\% & 2.5426 & 2.5682 & 2.6881 & 196608 & 97.96\% & 0.865 \\
\textit{MLP} & \textit{20} & \textit{4.5e-05} & \textit{0.00075} & \textit{31.4} & \textit{83.94\%} & \textit{2.5426} & \textit{2.5659} & \textit{2.6881} & \textit{196608} & \textit{99.24\%} & \textit{0.877} \\
MLP & 20 & 3e-05 & 0.001 & 39.5 & 83.12\% & 2.5426 & 2.5671 & 2.6881 & 196608 & 46.33\% & 0.924 \\
\textit{MLP} & \textit{20} & \textit{4e-05} & \textit{0.00075} & \textit{38.3} & \textit{85.18\%} & \textit{2.5426} & \textit{2.5641} & \textit{2.6881} & \textit{196608} & \textit{95.73\%} & \textit{0.889} \\
MLP & 20 & 3.5e-05 & 0.00075 & 43.2 & 84.11\% & 2.5426 & 2.5657 & 2.6881 & 196608 & 94.62\% & 0.874 \\
\textit{MLP} & \textit{20} & \textit{3e-05} & \textit{0.00075} & \textit{56.8} & \textit{87.23\%} & \textit{2.5426} & \textit{2.5612} & \textit{2.6881} & \textit{196608} & \textit{96.88\%} & \textit{0.894} \\
MLP & 20 & 2e-05 & 0.00075 & 68.1 & 84.18\% & 2.5426 & 2.5656 & 2.6881 & 196608 & 53.42\% & 0.898 \\
\hline
\end{tabular}
\caption{Gemma-7B Baseline SAEs (1024 sequence length) continued from \Cref{table:gemma_7b_baselines}.}
\label{table:gemma_7b_baselines2}
\end{table}

\begin{table}[ht!]
\centering
\scriptsize\begin{tabular}{|c|c|c|c|c|c|c|c|c|c|c|c|}
\hline
\textbf{\tiny{Site}} & \textbf{\tiny{Layer}} & \textbf{\tiny{Sparsity $\lambda$}} & \textbf{\tiny{LR}} & \textbf{\tiny{L0}} & \textbf{\tiny{\% CE Recovered}} & \textbf{\tiny{Clean CE Loss}} & \textbf{\tiny{SAE CE Loss}} & \textbf{\tiny{0 Abl. CE Loss}} & \textbf{\tiny{Width}} & \textbf{\tiny{\% Alive Features}} & \textbf{\tiny{Shrinkage $\gamma$}} \\
\hline
\textit{Resid} & \textit{6} & \textit{0.0012} & \textit{0.0003} & \textit{2.2} & \textit{95.55\%} & \textit{2.5426} & \textit{3.1483} & \textit{16.1549} & \textit{131072} & \textit{93.94\%} & \textit{1.006} \\
\textit{Resid} & \textit{6} & \textit{0.001} & \textit{0.0003} & \textit{3.0} & \textit{96.67\%} & \textit{2.5426} & \textit{2.9954} & \textit{16.1549} & \textit{131072} & \textit{96.24\%} & \textit{1.006} \\
\textit{Resid} & \textit{6} & \textit{0.0008} & \textit{0.0003} & \textit{4.3} & \textit{97.83\%} & \textit{2.5426} & \textit{2.8382} & \textit{16.1549} & \textit{131072} & \textit{97.52\%} & \textit{1.003} \\
\textit{Resid} & \textit{6} & \textit{0.0006} & \textit{0.0003} & \textit{7.0} & \textit{98.76\%} & \textit{2.5426} & \textit{2.7108} & \textit{16.1549} & \textit{131072} & \textit{98.3\%} & \textit{0.996} \\
\textit{Resid} & \textit{6} & \textit{0.0004} & \textit{0.0003} & \textit{14.3} & \textit{99.35\%} & \textit{2.5426} & \textit{2.6312} & \textit{16.1549} & \textit{131072} & \textit{98.68\%} & \textit{0.996} \\
\textit{Resid} & \textit{6} & \textit{0.0002} & \textit{0.0003} & \textit{45.9} & \textit{99.77\%} & \textit{2.5426} & \textit{2.5735} & \textit{16.1549} & \textit{131072} & \textit{99.51\%} & \textit{0.999} \\
Resid & 6 & 2e-05 & 0.0003 & 95.2 & 98.62\% & 2.5426 & 2.7302 & 16.1549 & 131072 & 45.13\% & 1.148 \\
\hline
\end{tabular}
\caption{Gemma-7B Gated SAEs (1024 sequence length). Continued in \Cref{table:gemma_7b_gated2}.}
\label{table:gemma_7b_gated}
\end{table}

\begin{table}[ht!]
\centering
\scriptsize\begin{tabular}{|c|c|c|c|c|c|c|c|c|c|c|c|}
\hline
\textbf{\tiny{Site}} & \textbf{\tiny{Layer}} & \textbf{\tiny{Sparsity $\lambda$}} & \textbf{\tiny{LR}} & \textbf{\tiny{L0}} & \textbf{\tiny{\% CE Recovered}} & \textbf{\tiny{Clean CE Loss}} & \textbf{\tiny{SAE CE Loss}} & \textbf{\tiny{0 Abl. CE Loss}} & \textbf{\tiny{Width}} & \textbf{\tiny{\% Alive Features}} & \textbf{\tiny{Shrinkage $\gamma$}} \\
\hline
\textit{MLP} & \textit{13} & \textit{5e-05} & \textit{0.0003} & \textit{196.4} & \textit{85.54\%} & \textit{2.5426} & \textit{2.5491} & \textit{2.5878} & \textit{131072} & \textit{76.56\%} & \textit{1.003} \\
\textit{MLP} & \textit{13} & \textit{3e-05} & \textit{0.0003} & \textit{766.5} & \textit{93.04\%} & \textit{2.5426} & \textit{2.5457} & \textit{2.5878} & \textit{131072} & \textit{86.81\%} & \textit{1.033} \\
\textit{MLP} & \textit{20} & \textit{0.00019} & \textit{0.0003} & \textit{24.4} & \textit{87.81\%} & \textit{2.5426} & \textit{2.5603} & \textit{2.6881} & \textit{131072} & \textit{99.91\%} & \textit{1.004} \\
\textit{MLP} & \textit{20} & \textit{0.00016} & \textit{0.0003} & \textit{32.7} & \textit{89.16\%} & \textit{2.5426} & \textit{2.5583} & \textit{2.6881} & \textit{131072} & \textit{99.94\%} & \textit{1.004} \\
\textit{MLP} & \textit{20} & \textit{0.00015} & \textit{0.0003} & \textit{36.4} & \textit{89.63\%} & \textit{2.5426} & \textit{2.5577} & \textit{2.6881} & \textit{131072} & \textit{99.95\%} & \textit{1.002} \\
\textit{MLP} & \textit{20} & \textit{0.00014} & \textit{0.0003} & \textit{40.8} & \textit{89.73\%} & \textit{2.5426} & \textit{2.5575} & \textit{2.6881} & \textit{131072} & \textit{99.96\%} & \textit{1.0} \\
\textit{MLP} & \textit{20} & \textit{0.00013} & \textit{0.0003} & \textit{46.6} & \textit{90.3\%} & \textit{2.5426} & \textit{2.5567} & \textit{2.6881} & \textit{131072} & \textit{99.95\%} & \textit{1.002} \\
\hline
\end{tabular}
\caption{Gemma-7B Gated SAEs (1024 sequence length). Continued from \Cref{table:gemma_7b_gated}}
\label{table:gemma_7b_gated2}
\end{table}

\begin{table}[ht!]
\centering
\scriptsize
\scriptsize\begin{tabular}{|c|c|c|c|c|c|c|c|c|c|c|c|}
\hline
\textbf{\tiny{Site}} & \textbf{\tiny{Layer}} & \textbf{\tiny{Sparsity $\lambda$}} & \textbf{\tiny{LR}} & \textbf{\tiny{L0}} & \textbf{\tiny{\% CE Recovered}} & \textbf{\tiny{Clean CE Loss}} & \textbf{\tiny{SAE CE Loss}} & \textbf{\tiny{0 Abl. CE Loss}} & \textbf{\tiny{Width}} & \textbf{\tiny{\% Alive Features}} & \textbf{\tiny{Shrinkage $\gamma$}} \\
\hline
\textit{Attn} & \textit{4} & \textit{8e-05} & \textit{0.001} & \textit{17.6} & \textit{81.04\%} & \textit{1.9699} & \textit{1.9824} & \textit{2.0361} & \textit{196608} & \textit{94.29\%} & \textit{0.827} \\
\textit{Attn} & \textit{4} & \textit{6e-05} & \textit{0.001} & \textit{24.2} & \textit{84.12\%} & \textit{1.9699} & \textit{1.9804} & \textit{2.0361} & \textit{196608} & \textit{95.76\%} & \textit{0.848} \\
\textit{Attn} & \textit{4} & \textit{3e-05} & \textit{0.001} & \textit{62.1} & \textit{90.96\%} & \textit{1.9699} & \textit{1.9759} & \textit{2.0361} & \textit{196608} & \textit{96.72\%} & \textit{0.93} \\
\textit{Attn} & \textit{12} & \textit{8e-05} & \textit{0.001} & \textit{16.1} & \textit{51.88\%} & \textit{1.9699} & \textit{1.9907} & \textit{2.0131} & \textit{196608} & \textit{65.73\%} & \textit{0.78} \\
\textit{Attn} & \textit{12} & \textit{6e-05} & \textit{0.001} & \textit{24.0} & \textit{58.46\%} & \textit{1.9699} & \textit{1.9878} & \textit{2.0131} & \textit{196608} & \textit{69.85\%} & \textit{0.802} \\
\textit{Attn} & \textit{12} & \textit{3e-05} & \textit{0.001} & \textit{75.0} & \textit{72.84\%} & \textit{1.9699} & \textit{1.9816} & \textit{2.0131} & \textit{196608} & \textit{73.04\%} & \textit{0.848} \\
\textit{Attn} & \textit{16} & \textit{0.00045} & \textit{0.001} & \textit{0.3} & \textit{-3.54\%} & \textit{1.9699} & \textit{2.0058} & \textit{2.0046} & \textit{49152} & \textit{20.1\%} & \textit{0.554} \\
\hline
\end{tabular}
\caption{Pythia-2.8B baseline SAEs (2048 sequence length). Continued in \Cref{table:pythia_2p8b_baseline2}.}
\label{table:pythia_2p8b_baseline}
\end{table}

\begin{table}[ht!]
\centering
\scriptsize
\scriptsize\begin{tabular}{|c|c|c|c|c|c|c|c|c|c|c|c|}
\hline
\textbf{\tiny{Site}} & \textbf{\tiny{Layer}} & \textbf{\tiny{Sparsity $\lambda$}} & \textbf{\tiny{LR}} & \textbf{\tiny{L0}} & \textbf{\tiny{\% CE Recovered}} & \textbf{\tiny{Clean CE Loss}} & \textbf{\tiny{SAE CE Loss}} & \textbf{\tiny{0 Abl. CE Loss}} & \textbf{\tiny{Width}} & \textbf{\tiny{\% Alive Features}} & \textbf{\tiny{Shrinkage $\gamma$}} \\
\hline
\textit{MLP} & \textit{12} & \textit{5e-05} & \textit{0.001} & \textit{28.2} & \textit{76.01\%} & \textit{1.9698} & \textit{1.9811} & \textit{2.0167} & \textit{196608} & \textit{99.45\%} & \textit{0.909} \\
\textit{MLP} & \textit{12} & \textit{7e-05} & \textit{0.001} & \textit{16.2} & \textit{71.94\%} & \textit{1.9698} & \textit{1.983} & \textit{2.0167} & \textit{196608} & \textit{99.14\%} & \textit{0.883} \\
\textit{MLP} & \textit{16} & \textit{2.5e-05} & \textit{0.001} & \textit{79.8} & \textit{78.44\%} & \textit{1.9698} & \textit{1.9785} & \textit{2.0098} & \textit{196608} & \textit{99.83\%} & \textit{0.919} \\
\textit{MLP} & \textit{16} & \textit{4e-05} & \textit{0.001} & \textit{29.0} & \textit{72.83\%} & \textit{1.9698} & \textit{1.9807} & \textit{2.0098} & \textit{196608} & \textit{99.82\%} & \textit{0.923} \\
\textit{MLP} & \textit{16} & \textit{3.5e-05} & \textit{0.001} & \textit{35.9} & \textit{73.95\%} & \textit{1.9698} & \textit{1.9803} & \textit{2.0098} & \textit{196608} & \textit{99.83\%} & \textit{0.914} \\
\textit{MLP} & \textit{16} & \textit{7.5e-05} & \textit{0.001} & \textit{11.2} & \textit{65.88\%} & \textit{1.9698} & \textit{1.9835} & \textit{2.0098} & \textit{196608} & \textit{99.45\%} & \textit{0.884} \\
\textit{MLP} & \textit{16} & \textit{4.5e-05} & \textit{0.001} & \textit{22.0} & \textit{70.73\%} & \textit{1.9698} & \textit{1.9815} & \textit{2.0098} & \textit{196608} & \textit{99.79\%} & \textit{0.901} \\
\hline
\end{tabular}
\caption{Pythia-2.8B baseline SAEs (2048 sequence length). Continued from \Cref{table:pythia_2p8b_baseline}.}
\label{table:pythia_2p8b_baseline2}
\end{table}

\begin{table}[ht!]
\centering
\scriptsize\begin{tabular}{|c|c|c|c|c|c|c|c|c|c|c|c|}
\hline
\textbf{\tiny{Site}} & \textbf{\tiny{Layer}} & \textbf{\tiny{Sparsity $\lambda$}} & \textbf{\tiny{LR}} & \textbf{\tiny{L0}} & \textbf{\tiny{\% CE Recovered}} & \textbf{\tiny{Clean CE Loss}} & \textbf{\tiny{SAE CE Loss}} & \textbf{\tiny{0 Abl. CE Loss}} & \textbf{\tiny{Width}} & \textbf{\tiny{\% Alive Features}} & \textbf{\tiny{Shrinkage $\gamma$}} \\
\hline
\textit{Attn} & \textit{4} & \textit{0.0006} & \textit{0.0003} & \textit{38.2} & \textit{92.85\%} & \textit{1.9699} & \textit{1.9746} & \textit{2.0361} & \textit{131072} & \textit{93.76\%} & \textit{1.006} \\
\textit{Attn} & \textit{4} & \textit{0.0004} & \textit{0.0003} & \textit{69.8} & \textit{94.82\%} & \textit{1.9699} & \textit{1.9733} & \textit{2.0361} & \textit{131072} & \textit{96.29\%} & \textit{1.0} \\
\textit{Attn} & \textit{4} & \textit{0.0008} & \textit{0.0003} & \textit{24.7} & \textit{90.94\%} & \textit{1.9699} & \textit{1.9759} & \textit{2.0361} & \textit{131072} & \textit{91.45\%} & \textit{1.007} \\
\textit{Attn} & \textit{12} & \textit{0.0006} & \textit{0.0003} & \textit{64.5} & \textit{82.04\%} & \textit{1.9699} & \textit{1.9776} & \textit{2.0131} & \textit{131072} & \textit{74.48\%} & \textit{0.99} \\
\textit{Attn} & \textit{12} & \textit{0.001} & \textit{0.0003} & \textit{27.1} & \textit{73.09\%} & \textit{1.9699} & \textit{1.9815} & \textit{2.0131} & \textit{131072} & \textit{63.68\%} & \textit{0.987} \\
\textit{Attn} & \textit{12} & \textit{0.0008} & \textit{0.0003} & \textit{40.5} & \textit{77.52\%} & \textit{1.9699} & \textit{1.9796} & \textit{2.0131} & \textit{131072} & \textit{67.74\%} & \textit{0.998} \\
\textit{Attn} & \textit{16} & \textit{0.001} & \textit{0.0003} & \textit{17.2} & \textit{79.67\%} & \textit{1.9699} & \textit{1.9769} & \textit{2.0046} & \textit{32768} & \textit{89.76\%} & \textit{0.988} \\
\hline
\end{tabular}
\caption{Pythia-2.8B Gated SAEs (2048 sequence length). Continued in \Cref{table:pythia_2p8b_gated2}.}
\label{table:pythia_2p8b_gated}
\end{table}

\begin{table}[ht!]
\centering
\scriptsize\begin{tabular}{|c|c|c|c|c|c|c|c|c|c|c|c|}
\hline
\textbf{\tiny{Site}} & \textbf{\tiny{Layer}} & \textbf{\tiny{Sparsity $\lambda$}} & \textbf{\tiny{LR}} & \textbf{\tiny{L0}} & \textbf{\tiny{\% CE Recovered}} & \textbf{\tiny{Clean CE Loss}} & \textbf{\tiny{Spliced SAE CE Loss}} & \textbf{\tiny{Zero Ablation CE Loss}} & \textbf{\tiny{Shrinkage $\gamma$}} \\
\hline
\textit{MLP} & \textit{16} & \textit{0.0008} & \textit{0.0003} & \textit{51.0} & \textit{80.32\%} & \textit{1.9698} & \textit{1.9777} & \textit{2.0098} & \textit{131072} & \textit{99.05\%} & \textit{1.002} \\
\textit{MLP} & \textit{16} & \textit{0.0016} & \textit{0.0003} & \textit{12.4} & \textit{70.76\%} & \textit{1.9698} & \textit{1.9815} & \textit{2.0098} & \textit{131072} & \textit{97.38\%} & \textit{1.005} \\
\textit{MLP} & \textit{16} & \textit{0.0007} & \textit{0.0003} & \textit{70.1} & \textit{82.09\%} & \textit{1.9698} & \textit{1.977} & \textit{2.0098} & \textit{131072} & \textit{99.32\%} & \textit{1.001} \\
\textit{MLP} & \textit{16} & \textit{0.0014} & \textit{0.0003} & \textit{16.1} & \textit{72.62\%} & \textit{1.9698} & \textit{1.9808} & \textit{2.0098} & \textit{131072} & \textit{97.48\%} & \textit{1.007} \\
\textit{MLP} & \textit{16} & \textit{0.0012} & \textit{0.0003} & \textit{21.9} & \textit{75.12\%} & \textit{1.9698} & \textit{1.9798} & \textit{2.0098} & \textit{131072} & \textit{98.18\%} & \textit{1.012} \\
\textit{MLP} & \textit{16} & \textit{0.0009} & \textit{0.0003} & \textit{38.3} & \textit{78.41\%} & \textit{1.9698} & \textit{1.9785} & \textit{2.0098} & \textit{131072} & \textit{98.72\%} & \textit{0.993} \\
\textit{MLP} & \textit{20} & \textit{0.0008} & \textit{0.0003} & \textit{51.0} & \textit{94.28\%} & \textit{1.9698} & \textit{1.9774} & \textit{2.1022} & \textit{131072} & \textit{99.06\%} & \textit{1.007} \\
\hline
\end{tabular}
\caption{Pythia-2.8B Gated SAEs (2048 sequence length). Continued from \Cref{table:pythia_2p8b_gated}.}
\label{table:pythia_2p8b_gated2}
\end{table}

\section{Equivalence between gated encoder with tied weights and linear encoder with non-standard activation function}
\label{app:gated_equiv}

\newcommand{\wsgate}{\mathbf{w}_\text{gate}}
\newcommand{\wsmag}{\mathbf{w}_\text{mag}}
\newcommand{\bsgate}{b_\text{gate}}
\newcommand{\bsmag}{b_\text{mag}}
\newcommand{\rsmag}{r_\text{mag}}
\newcommand{\exprsmag}{\rho_\text{mag}}

In this section we show under the weight sharing scheme defined in \cref{eq:tying_scheme}, a gated encoder as defined in \cref{eq:gated_encoder} is equivalent to a linear layer with a non-standard (and parameterised) activation function.

Without loss of generality, consider the case of a single latent feature ($M = 1$) and set the pre-encoder bias to zero. In this case, the gated encoder is defined as
\begin{equation}
    \tilde{f}(\mathbf{x}) := \mathbb{1}_{\wsgate \cdot \mathbf{x}  + \bsgate > \mathbf{0}} \;\text{ReLU}\left(\wsmag \cdot \mathbf{x} + \bsmag\right)
\label{eq:gated_encoder_scalar}
\end{equation}
and the weight sharing scheme becomes
\begin{equation}
\wsmag := \exprsmag \wsgate
\label{eq:tying_sceme_scalar}
\end{equation}
with a non-negative parameter $\exprsmag\equiv\exp(\rmag)$.

Substituting \cref{eq:tying_sceme_scalar} into \cref{eq:gated_encoder_scalar} and re-arranging, we can re-express $\tilde{f}(\mathbf{x})$ as a single linear layer
\begin{equation}
\tilde{f}(\mathbf{x}) := \sigma_{\bsmag-\exprsmag\bsgate}\left(\wsmag \cdot \mathbf{x} + \bsmag\right)
\label{eq:gated_encoder_scalar_alt}
\end{equation}
with the parameterised activation function
\begin{equation}
\sigma_\theta(z) := \mathbb{1}_{z > \theta}\; \text{ReLU}\left(z\right).
\label{eq:equivalent_act_fn}
\end{equation}
called JumpReLU in a different context \citep{erichson2019jumprelu}. \cref{fig:jump_relu} illustrates the shape of this activation function.

\section{A toy setting where Jump ReLU SAEs outperform baseline SAEs}
\label{app:toy-model}

\begin{figure}[htp]
\centering
\begin{subfigure}{.5\textwidth}
  \centering
  \includegraphics[width=\linewidth]{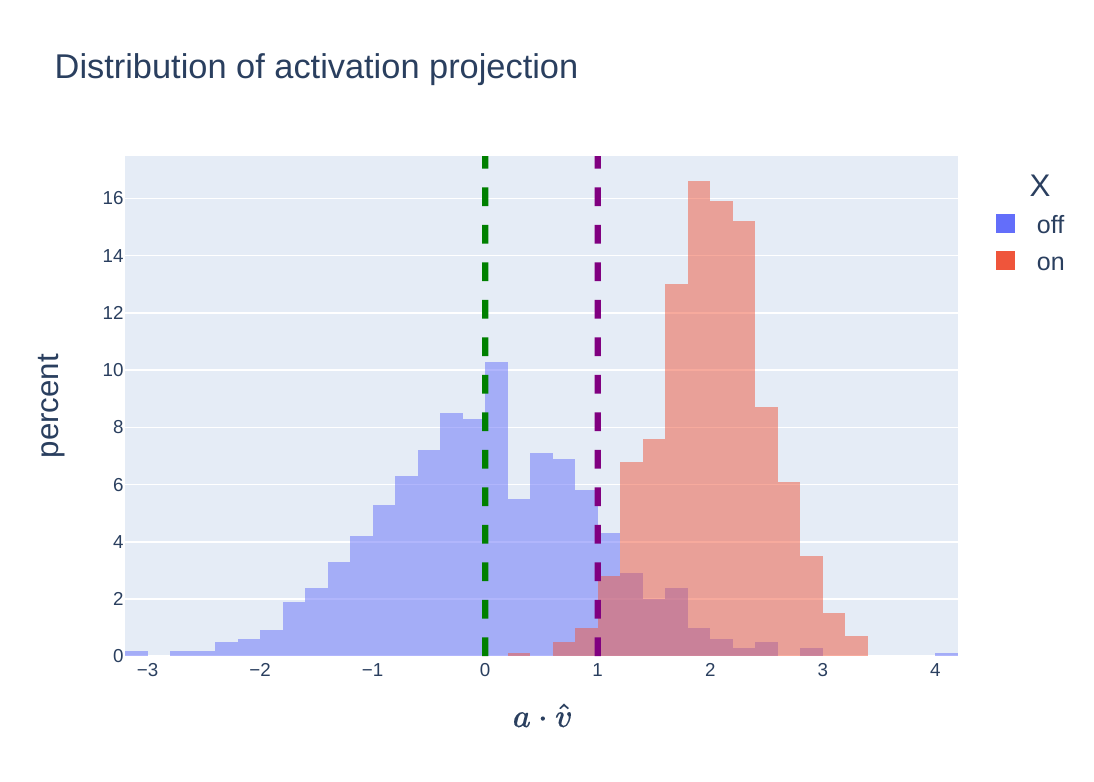}
  \caption{Empirical distribution of $a\cdot\hat{v}$}
  \label{fig:toy_model_hist}
\end{subfigure}%
\hfill
\begin{subfigure}{.5\textwidth}
  \centering
  \includegraphics[width=\linewidth]{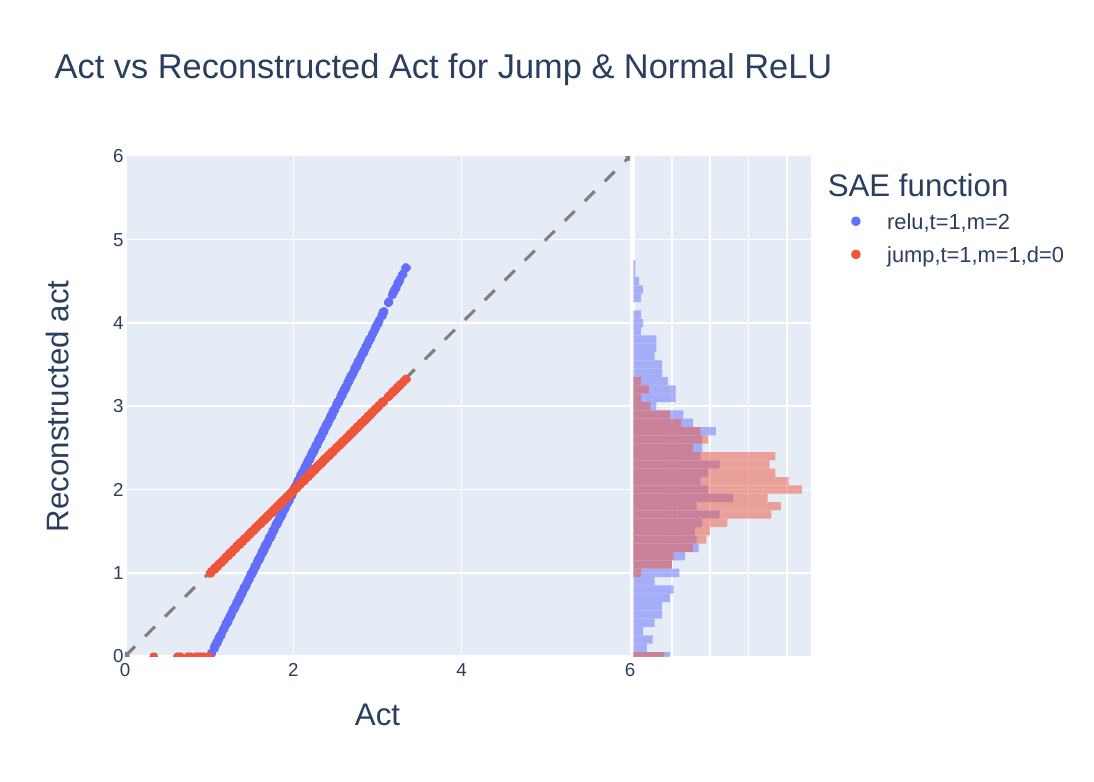}
  \caption{Comparison of $a\cdot\hat{v}$ and its reconstruction for a standard ReLU (blue) and a Jump ReLU (red)}
  \label{fig:toy_model_scatter}
\end{subfigure}%
\caption{(a) Shows the empirical distribution of $a\cdot\hat{v}$ in the toy model where $a \cdot \hat{v} \sim \mathbb{1}_{X \textrm{ is on}}(0.5 Z_1 + 2) + \mathbb{1}_{X \textrm{ is off}} Z_2$. The green line is $x=0$, the purple line is $x=1$. (b) shows a scatter plot of the reconstruction of $a\cdot \hat{v}$ against $a\cdot\hat{v}$ for two possible SAE activations: the blue line is a standard ReLU (with $t=1,m=2$), i.e. setting a threshold at the purple line and then taking twice the distance from it, and the red line is a Jump ReLU (with $t=1,m=1,d=0$), i.e. setting a threshold at the purple line and then taking the distance from the \textit{green} line. Note that the Jump ReLU gives a perfect reconstruction (above one) while the standard ReLU is highly imperfect.}
\label{fig:toy_model}
\end{figure}

An additional reason that Gated SAEs may perform baseline SAEs, beyond resolving shrinkage, is that they are a more expressive architecture: at inference time, they're equivalent to an SAE with the ReLU replaced by a potentially discontinuous Jump ReLU \citep{erichson2019jumprelu}, as shown in \cref{app:gated_equiv}. In this appendix we present a toy setting where a Jump ReLU is a more natural activation function for sparsely reconstructing activations than a ReLU. We adopt a more intuitive and less formal style, for pedagogical purposes. 

Consider a sparsely activating but continuously valued feature $X$, and a fixed unit encoder direction $\vhat$. If $X$ is off ($X=0$), the projection of activations $\mathbf{a}$ onto $\vhat$ is normally distributed as $\mathcal{N}(0, 1)$ (simulating noise from non-orthogonal features firing); if $X$ is on ($X>0$) then the projection is normally distributed as $\mathcal{N}(2, 1/4)$. Suppose further that $X$ is on with $50\%$ probability. So $a \cdot \hat{v} \sim \mathbb{1}_{X \textrm{ is on}}(0.5 Z_1 + 2) + \mathbb{1}_{X \textrm{ is off}} Z_2$. Where $A$ is the activation and $Z_1, Z_2 \sim \mathcal{N}(0, 1^2)$ are standard 1D Gaussians. The empirical distribution is shown in Figure \ref{fig:toy_model_hist}.

Consider the problem of fitting a ReLU encoder to this. With fixed encoder unit direction $\hat{v}$, the encoder is parametrised by a bias $b$ and magnitude $m$, where $a \to \max(m a \cdot \hat{v} + b, 0)$. $b$ can be reparametrised in terms of a threshold $t = \frac{-b}{m}$, so it is now $a \to \mathbb{1}_{a\cdot \hat{v} > t} m(a\cdot \hat{v}-t)$. Geometrically, we set some threshold $t$, a vertical line. Everything to the left is set to zero, and everything to the right is set to some multiple of \textit{the distance from the line}. 

This illuminates the core problem with ReLU SAEs: the threshold both determines whether to fire at all, and gives an origin to take the distance from if firing. The optimal reconstruction when $X$ is on requires us to take $t=0$. But now we fire half the time when $X$ is off, as a lot of the blue histogram is to the right of the green line. However, if we take a high enough threshold to exclude most blue, e.g. $t=1$, we now need to take the distance from $t=1$ too when $X$ is on, distorting things, even if we try to correct by rescaling with $m$, see the blue line in Figure \ref{fig:toy_model_scatter}. 

Jump ReLUs solve this problem. Mathematically, we can parametrise a Jump ReLU, at least in the setting of Gated SAEs, as $\mathbb{1}_{(x>t)} m(x-d)$. Geometrically, we now have two vertical lines. $x=t$ sets the threshold: anything to the left is set to zero. $x=d$ sets the origin point (for some $d\leq t$), we return the distance to $d$ times some magnitude $m$. This solves our problem: we can set $t=1$ (the purple line), $d=0$ (the green line) and $m=1$ (no distortion correction needed), and get the blue line in Figure \ref{fig:toy_model_scatter}, a near perfect reconstruction!

Some caveats and reflections on this toy model:
\begin{compactitem}
    \item The numbers $t=1$, $m=2$ are likely not the mathematically optimal solution, and are given for pedagogical purposes, but this seems unlikely to change the conceptual takeaways.
    \item This toy model has not been empirically tested, and could be totally off. But we've found it useful for building intuition.
    \item Why was it realistic to assume that the projection wasn't just zero when $X$ was off? Because there are likely many other non-orthogonal features firing, due to superposition, which in aggregate creates significant interference. Indeed, a common problem when studying SAE features/other interpretable directions is that, while the tails are monosemantic, activations near zero are very noisy (see e.g. the Arabic feature in \citet{bricken2023monosemanticity} or the sentiment feature in \citet{tigges2023linear}). We speculate that this is a consequence of ReLU SAEs needing to choose a threshold with a mix of on and off activations (a mix of red and blue in \ref{fig:toy_model_hist}) to minimise distortion to the tails, as L1 does not penalise incorrectly firing at small magnitudes much. We hope that Gated SAEs may have less of these issues, as they can just have a large gap between $t$ and $d$.
    \item We asserted that X was a sparsely activating but continuous feature. It's an open question how much such features actually exist in models (though at least some likely do \citep{gurnee2024language}). Our intuition is that most features are essentially binary (e.g. "is this about basketball"), but that models track their confidences in them as coefficients of the feature directions, and that reconstructing the precise coefficients matters (otherwise, we should just discretise SAE activations at inference time!), so they can be thought of as continuous. We think understanding this better is a promising direction of future work.
    \item In real models, the probability that $X$ fires is likely much less than $50\%$! But this assumption simplified the reasoning and diagrams, without qualitatively changing much. 
    \item We assumed that the encoder direction ($\hat{v}$) was frozen, even if its magnitude was not. This was a simplifying assumption that is clearly false in Gated SAEs, and indeed, as shown in Section \ref{subsec:rescale_expt}, their ability to choose different directions to a standard SAE is key to performance, and they outperform a standard SAE fine-tuned with Jump ReLUs. 
    \item The main reason a standard ReLU SAE doesn't want $t=0$ is that this includes too many activations when $X$ is off. But, actually, this is good for reconstruction, just bad for L1 and sparsity. Gated SAEs decouple L1 from their encoder directions, making it hard to reason clearly about whether the need for a high threshold would still apply in a hypothetical Gated SAE with standard ReLUs (though in an actual Gated SAE, the L1 crucially \textit{is} still applied to $t$)
\end{compactitem}

\section{Pseudo-code for Gated SAEs and the Gated SAE loss function}
\label{app:pseudocode}
\begin{figure}[H]
\begin{lstlisting}
def gated_sae(x, W_gate, b_gate, W_mag, b_mag, W_dec, b_dec):
  # Apply pre-encoder bias
  x_center = x - b_dec

  # Gating encoder (estimates which features are active)
  active_features = ((x_center @ W_gate + b_gate) > 0)

  # Magnitudes encoder (estimates active features' magnitudes)
  feature_magnitudes = relu(x_center @ W_mag + b_mag)

  # Multiply both before decoding
  return (active_features * feature_magnitudes) @ W_dec + b_dec
\end{lstlisting}
\caption{Pseudo-code for the Gated SAE forward pass.}
\label{code:gated_sae}
\end{figure}

\begin{figure}[H]
\begin{lstlisting}
def loss(x, W_gate, b_gate, W_mag, b_mag, W_dec, b_dec):
  gated_sae_loss = 0.0

  # We'll use the reconstruction from the baseline forward pass to train
  # the magnitudes encoder and decoder. Note we don't apply any sparsity
  # penalty here. Also, no gradient will propagate back to W_gate or b_gate
  # due to binarising the gated activations to zero or one.
  reconstruction = gated_sae(x, W_gate, b_gate, W_mag, b_mag, W_dec, b_dec)
  gated_sae_loss += sum((reconstruction - x)**2, axis=-1)

  # We apply a L1 penalty on the gated encoder activations (pre-binarising,
  # post-ReLU) to incentivise them to be sparse
  x_center = x - b_dec
  via_gate_feature_magnitudes = relu(x_center @ W_gate + b_gate)
  gated_sae_loss += l1_coef * sum(via_gate_feature_magnitudes, axis=-1)

  # Currently the gated encoder only has gradient signal to be sparse, and
  # not to reconstruct well, so we also do a "via gate" reconstruction, to
  # give it an appropriate gradient signal. We stop the gradients to the
  # decoder parameters in this forward pass, as we don't want these to be
  # influenced by this auxiliary task.
  via_gate_reconstruction = (
    via_gate_feature_magnitudes @ stop_gradient(W_dec)
    + stop_gradient(b_dec)
  )
  gated_sae_loss += sum((via_gate_reconstruction - x)**2, axis=-1)

  return gated_sae_loss
\end{lstlisting}
    \caption{Pseudo-code for the Gated SAE loss function. Note that this pseudo-code is written for expositional clarity. In practice, taking into account parameter tying, it would be more efficient to rearrange the computation to avoid unnecessarily duplicated operations.}
    \label{code:gated_loss}
\end{figure}

\section{Further analysis of the human interpretability study}
\label{app:study}

We perform some further analysis on the data from \Cref{subsec:manual_interp}, to understand the impact of different sites, layers, and raters.

\subsection{Sites}

We first pose the question of whether there's evidence that the sites had different interpretability outcomes. A Friedman test across sites shows significant differences (at $p=.047$) between the Gated-vs-Baseline differences, though not ($p=.92$) between the raw labels.

Breaking down by site and repeating the Wilcoxon-Pratt one-sided tests and computing confidence intervals, we find the result on MLP outputs is strongest, with mean .40, significance $p=.003$, and CI [.18, .63]; this is as compared with the attention outputs ($p=.47$, mean .05, CI [-.16, .26]) and final residual ($p=.59$, mean -0.07, CI [-.28, .12]) SAEs.

\subsection{Layers}

Next we test whether different layers had different outcomes. We do this separately for the 2 models, since the layers aren't directly comparable. We run 2 tests in each setting: Page's trend test (which tests for a monotone trend across layers) and the Friedman test (which tests for any difference, without any expectation of a monotone trend).

\begin{figure}
    \centering
    \includegraphics[width=\textwidth]{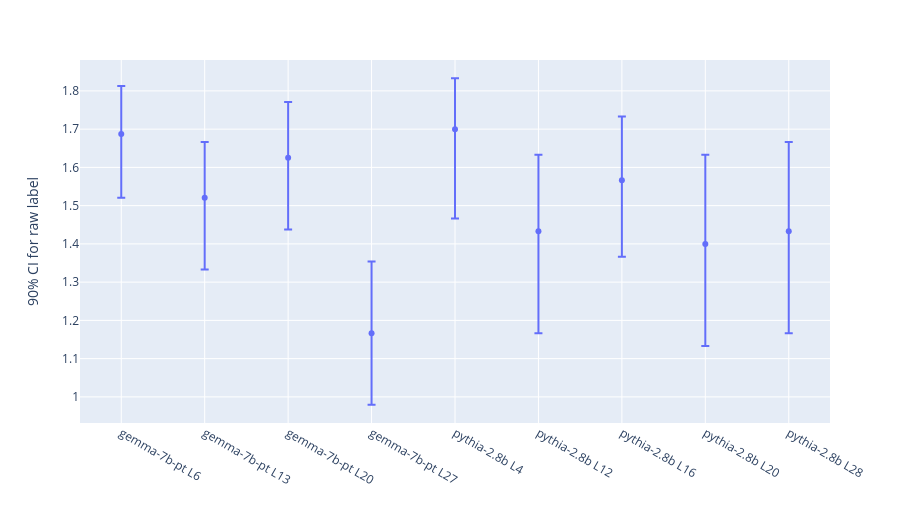}
    \caption{Per-layer 90\% confidence intervals for the mean interpretability label}
    \label{fig:label_cis_raw}
\end{figure}

\begin{figure}
    \centering
    \includegraphics[width=\textwidth]{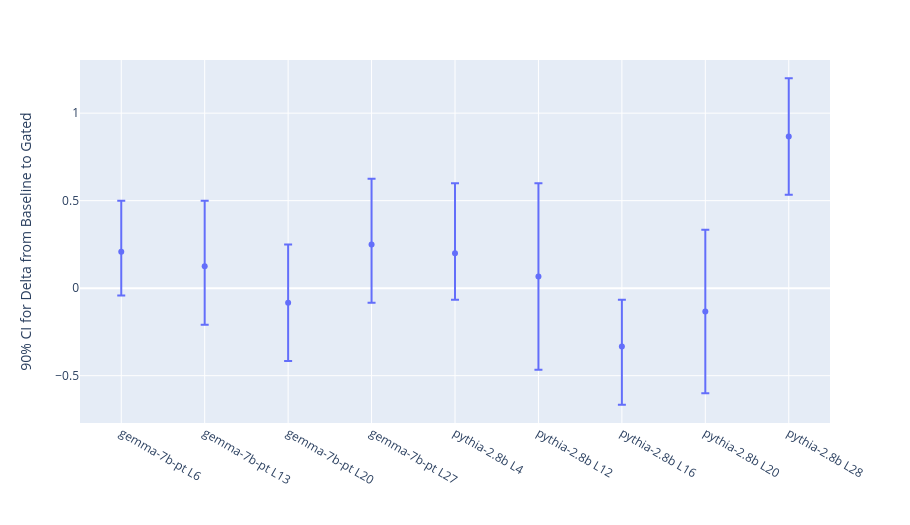}
    \caption{Per-layer 90\% confidence intervals for the Gated-vs-Baseline label difference}
    \label{fig:label_cis_delta}
\end{figure}

Results are presented in \Cref{tbl:layer_tests}; they suggest there are some significant nonmonotone differences between layers. To elucidate this, we present 90\% BCa bootstrap confidence intervals of the mean raw label (where `No'=0, `Maybe'=1, `Yes'=2) and the Gated-vs-Baseline difference, per layer, in \Cref{fig:label_cis_raw} and \Cref{fig:label_cis_delta}, respectively.

\begin{table}
\begin{tabular}{ |c|l|l| } 
 \hline
$p$-values & Raw label & Delta from Baseline to Gated \\ 
 \hline
Pythia-2.8B (Page's trend test) & .50 & .13 \\ 
Pythia-2.8B (Friedman test) & .57 & .05 \\ 
Gemma-7B (Page's trend test) & .37 & .31 \\
Gemma-7B (Friedman test) & .003 & .64 \\
 \hline
\end{tabular}
\caption{Layer significance tests}
\label{tbl:layer_tests}
\end{table}



\begin{figure}
    \centering
    \includegraphics[width=.65\textwidth]{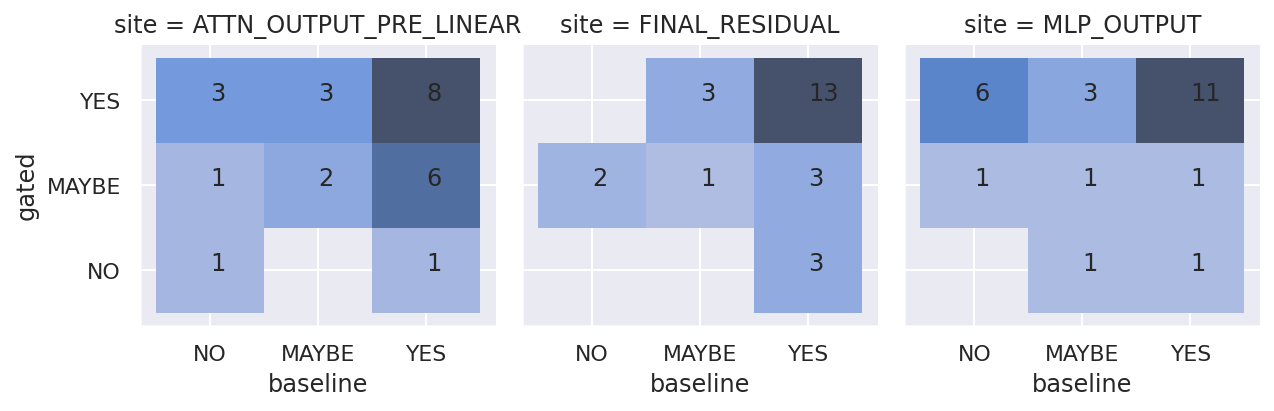}
    \includegraphics[width=\textwidth]{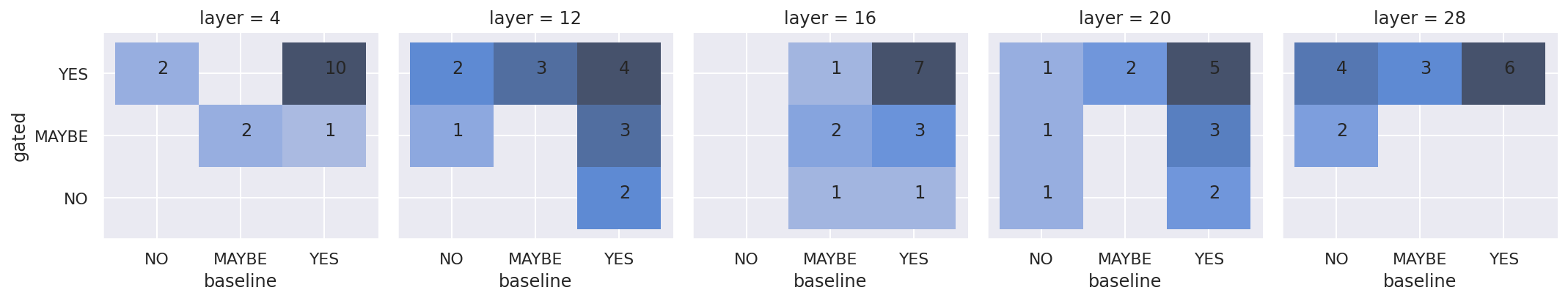}
    \caption{Contingency tables for the paired (gated vs baseline) interpretability labels, for Pythia-2.8B}
    \label{fig:contingency_pythia}
\end{figure}
\begin{figure}
    \centering
    \includegraphics[width=.55\textwidth]{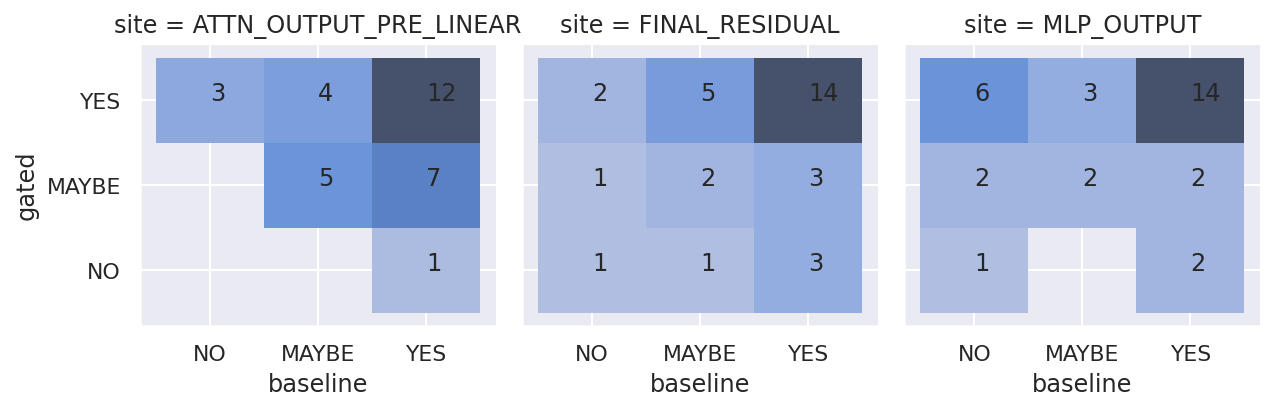}
    \includegraphics[width=\textwidth]{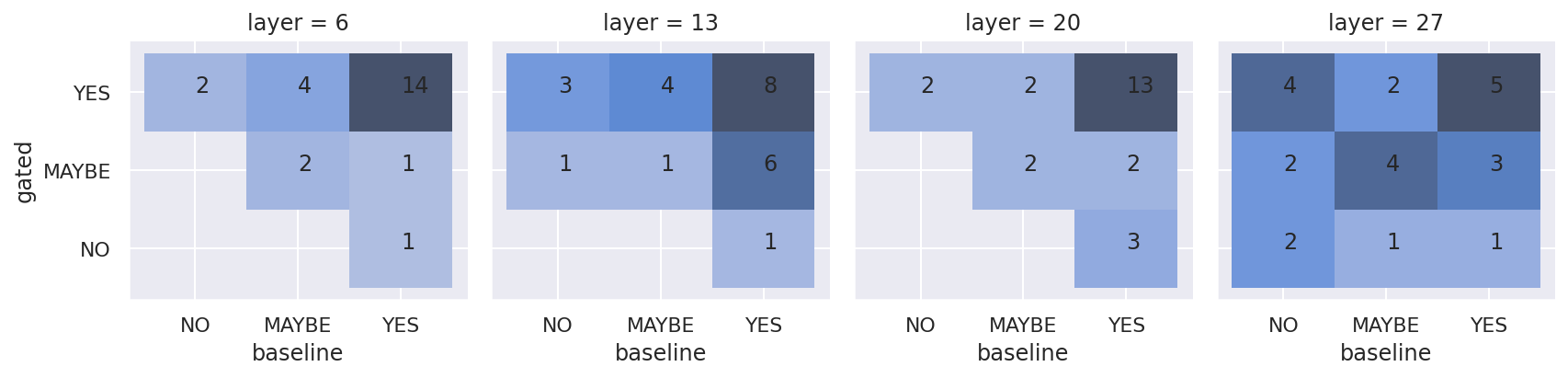}
    \caption{Contingency tables for the paired (gated vs baseline) interpretability labels, for Gemma-7B}
    \label{fig:contingency_gemma}
\end{figure}

\subsection{Raters}

In \Cref{tbl:rater_tests} we present test results weakly suggesting that the raters differed in their judgments. This underscores that there's still a significant subjective component to this interpretability labeling. (Notably, different raters saw different proportions of Pythia vs Gemma features, so aggregating across the models is partially confounded by that.)

\begin{table}
\begin{tabular}{ |c|l|l| } 
 \hline
$p$-values & Raw label & Delta from Baseline to Gated \\ 
 \hline
Across models (Kruskal-Wallis H-test)  & .01 & .71 \\ 
Pythia-2.8B (Friedman test) & .13 & .05 \\ 
Gemma-7B (Friedman test) & .03 & .76 \\
 \hline
\end{tabular}
\caption{Rater significance tests}
\label{tbl:rater_tests}
\end{table}

\end{document}